# Latent Diffusion-GAN: Adversarial Learning in the Autoencoded Latent Space

U-Chae Jun, Jaeeun Ko, and Jiwoo Kang†

Sookmyung Women's University, South Korea
{wjsdbco, rhwodms1223, jwkang}@sookmyung.ac.kr

**Abstract**

*Diffusion models are powerful generative frameworks for producing high-quality images by denoising latent variables from random noise. However, training with likelihood-based objectives, such as denoising score matching, can lead to locally over-smoothed high-frequency details, including fine textures and sharp edges, thereby limiting perceptual fidelity and structural detail. Adversarial training with GANs enhances sharpness but typically requires additional discriminator networks, increasing computational costs and destabilizing training. To this end, we propose Latent Diffusion Generative Adversarial Networks (LD-GAN), a novel framework that seamlessly integrates adversarial learning into diffusion models without modifying their original pipeline. LD-GAN leverages the pretrained variational autoencoder (VAE) in latent diffusion models as an energy-based discriminator, enabling adversarial training without extra parameters and preserving the structured latent priors learned from large datasets. We also introduce a structural consistency energy that aligns encoder and decoder feature representations, thereby enhancing perceptual quality and compatibility with the pretrained latent space. Extensive experiments demonstrate that LD-GAN significantly improves sample fidelity, perceptual sharpness, and diversity over state-of-the-art baseline methods across various generation tasks while ensuring efficient training dynamics.*

**CCS Concepts**

• ***Computing methodologies*** → *Computer vision;*

## 1 Introduction

Generative modeling has progressed significantly, with diffusion models [HJA20; SWMG15] emerging as a powerful framework for high-quality image synthesis. Diffusion models iteratively refine noisy inputs through denoising steps, achieving remarkable results in text-to-image generation [JKK*25; RBL*22] and video synthesis [SPH*23]. Despite their impressive generative capabilities, diffusion models primarily optimize likelihood-based objectives for statistical realism but often lack perceptual sharpness and structural coherence. It often leads to blurry textures, loss of fine details, and suboptimal performance in high-resolution synthesis tasks [DN21; VKK21].

Generative Adversarial Networks (GANs) [GPM*14], in contrast, are well known for generating sharp and realistic images by leveraging a discriminator to ensure perceptual fidelity. Recent efforts have integrated diffusion models with adversarial training, combining the structured sampling of diffusion models with the high-fidelity generation of GANs [WZH*23; SHCS22; NGH*22]. Diffusion-GAN [WZH*23] integrates adversarial loss into the diffusion framework, improving sample fidelity but facing challenges in stabilizing adversarial optimization. Adversarial diffusion distillation [SHCS22] distills knowledge from a diffusion model into a compact adversarially trained generator, while adversarial purification [NGH*22] applies diffusion models to remove adversarial perturbations. Despite these advances, adversarial training in diffusion models remains challenging due to instability, additional discriminator requirements, and potential conflicts between likelihood-based objectives and adversarial constraints.

To address these limitations, we propose *Latent Diffusion Generative Adversarial Networks (LD-GAN)*, a novel framework that integrates adversarial learning into diffusion models while preserving their training pipeline. Unlike previous approaches that require an additional discriminator network, LD-GAN *leverages the variational autoencoder (VAE) structure* within the latent diffusion model as an *energy-based discriminator*. Inspired by Energy-Based GANs (EB-GANs) [ZML17], we reformulate the discriminator as an energy function that assesses the consistency between latent codes and reconstructions generated by the VAE's encoder and decoder. The proposed architecture is illustrated in Fig. 1. Specifically, LD-GAN introduces an adversarial objective in the latent space, where positive (*i.e.*, real) samples are drawn from intermediate noisy latents $z_t \sim \mathcal{U}(0,T)$, and negative (*i.e.*, fake) samples are obtained by partially denoising noise-corrupted latents

† Corresponding author.

(*e.g.*, $\hat{z}_t = \text{UNet}(z_t + \epsilon)$). The representations are assessed using an energy-based discriminator composed of the VAE decoder and encoder in the latent diffusion model, evaluating reconstruction consistency in latent and feature spaces. This formulation enables adversarial training *without modifying the original diffusion pipeline*, and the pretrained VAE components allow LD-GAN to inherit the structured latent prior learned from large-scale datasets, enhancing training stability and sample fidelity.

A key challenge in adversarially training diffusion models is balancing adversarial loss with the likelihood-based denoising objective. Traditional adversarial training can introduce instability, often leading to mode collapse or loss of structural fidelity [DN21; SME21]. Direct adversarial training can cause the diffusion model to diverge from its structured generative prior, as the discriminator and generator optimize competing objectives. To address this, we introduce a *structural consistency energy* that enforces multi-scale alignment between encoder and decoder features in the latent diffusion model. This regularization prevents adversarial updates from distorting latent structure and ensures that decoder fine-tuning remains compatible with the pre-trained VAE prior. Our structural consistency energy is self-contained, unlike conventional perceptual losses [ZIE*18], allowing for lightweight and effective perceptual regularization.

Our main contributions are: (1) We introduce *LD-GAN*, an adversarially trained diffusion model without an additional discriminator, reducing computational overhead while maintaining the generative prior; (2) We propose an *energy-based discriminator* that repurposes the pretrained VAE encoder and decoder of the latent diffusion to evaluate consistency between latent representations and reconstructions; (3) We introduce a *structural consistency energy* that stabilizes training by aligning encoder and decoder features, preserving structured priors, and improving perceptual quality; (4) Our method leverages the pretrained latent structures from large-scale training to enable stable adversarial learning in the latent space; (5) We demonstrate that *LD-GAN* enhances perceptual sharpness, structural coherence, and sample diversity in generative tasks, including text-to-image generation, conditional text-to-image generation, and 2D-to-3D synthesis, establishing a robust and generalizable adversarial framework for diffusion models.

## 2 Related Works

### 2.1 Generative Adversarial Networks (GANs)

Generative Adversarial Networks (GANs) [GPM*14] have significantly advanced generative modeling by introducing an adversarial framework in which a generator $G$ synthesizes realistic samples, and a discriminator $D$ distinguishes real from generated samples. Traditional GANs employ a classifier-based discriminator, whereas Energy-Based Generative Adversarial Networks (EB-GANs) [ZML17] reformulate the discriminator as an energy function, establishing a structured separation between real and generated samples.

*Energy-Based Formulation.* In EB-GAN, the discriminator $D(x)$ assigns energy values to inputs $x$, encouraging lower energy for real samples than generated ones. The discriminator is trained:

$$L_D = D(x) + \max(0, m - D(G(z))), \tag{1}$$

where $m$ is a positive margin enforcing separation between real and fake energies. The generator minimizes the discriminator's energy for generated samples:

$$L_G = D(G(z)). \tag{2}$$

*Advancements in Energy-Based Generative Models.* Recent work addressed EB-GAN's limitations with self-supervised and contrastive techniques. [SWCM21] combined contrastive learning with energy-based models for out-of-distribution detection while [TMJS20] employed self-supervised learning to improve anomaly detection. However, EB-GAN variants are restricted to the image domain, limiting adaptability to structured latent spaces. Moreover, instability in energy-based training can arise from overfitting in the energy function [DM19], indicating the need for further improvements.

*Relation to Our Work.* We extend EB-GANs with autoencoding-based energy formulation in the latent space instead of the image domain, utilizing a *VAE as an energy-based discriminator*. Our discriminator leverages autoencoding priors, providing structured adversarial guidance over conventional pixel-based discrimination.

### 2.2 Diffusion Models with Adversarial Training

Denoising diffusion probabilistic models (DDPMs) [HJA20] are generative models that add noise in a forward process and learn to denoise in a reverse process. Extensions like DDIMs [SME21] and improved diffusion models [KAAL22; ND21] enhance sampling efficiency and image quality. Score-based methods, such as NCSNs [SE19], leverage Langevin dynamics, while Latent Diffusion Models (LDMs) [RBL*22] operate in a learned latent space to lower computational costs.

*Integrating Diffusion Models and Adversarial Approaches.* Diffusion models effectively cover data distributions but suffer from slow sampling and overly smooth textures due to mean-squared error objectives [VKK21; RBL*22]. GANs, in contrast, generate sharper details and offer faster inference but can experience mode collapse [GPM*14]. Combining the broad coverage of diffusion models with the high-fidelity detail of adversarial training is therefore an appealing strategy. Recent studies show that synthetic data from diffusion models can strengthen adversarial robustness [WPD*23] and facilitate adversarial purification [NGH*22]. Furthermore, progressive distillation [SHCS22] transfers diffusion strengths into a GAN framework, and Diffusion-GAN [WZH*23] embeds an adversarial loss within diffusion, although stabilizing the combined objectives remains challenging. While some recent approaches [XKV22; XZXH24; KZB*24; YGZ*24] leverage adversarial objectives to accelerate diffusion inference through distillation or reduced-step sampling. The slow iterative process of diffusion sampling motivates such research, but aggressive reduction of sampling steps often degrades generation quality.

*Challenges in Adversarially Training Diffusion Models.* Merging adversarial and diffusion training, despite promising results, presents several challenges. First, Min-Max GAN objective may conflict with the likelihood-based diffusion, causing unstable optimization [DN21; SME21]. Second, suitable discriminator selection is challenging; pixel-space discrimination is common yet inefficient due to diffusion's structured latent manifolds [XKV22].

Third, adversarial objectives risk mode collapse, reducing sample diversity [HJA20; SME21]. Moreover, the computational demands of diffusion models require carefully designed noise schedules and architectures for efficient training and sampling.

*Improving Diffusion Models with Latent-Space Adversarial Learning.* A promising solution is adversarial training in a learned latent space. Our framework stabilizes the diffusion training process in a structured latent domain by incorporating a VAE or similar architecture, reducing conflicts between adversarial and likelihood objectives. This strategy enhances perceptual quality through adversarial guidance while maintaining the broad coverage of diffusion-based methods. Since the proposed method operates purely at training time and does not modify the diffusion sampling procedure, it is compatible with existing inference acceleration techniques, such as fast samplers [WCHN22; LZB*25] or diffusion distillation [KZB*24; YGZ*24].

## 3 Latent Diffusion GAN (LD-GAN)

### 3.1 Preliminaries

Our approach builds upon two key components: (i) *denoising loss in latent diffusion models* for likelihood-based training, and (ii) *energy-based adversarial learning* for structured discrimination.

*Denoising Loss in Latent Diffusion Models.* Denoising diffusion probabilistic models (DDPMs) [HJA20] define a generative process by iteratively refining noisy samples. Given an image $x$, a variational autoencoder (VAE) [KW14] encodes it into a latent representation $z_0 = \mathrm{Enc}(x)$. The forward diffusion process applies a Markovian noising:

$$q(z_t|z_{t-1}) = \mathcal{N}(z_t; \sqrt{\alpha_t} z_{t-1}, (1-\alpha_t)I), \tag{3}$$

where $\alpha_t$ controls the noise accumulation rate. The reverse denoising process is parameterized by a neural network $\epsilon_\theta(z_t, t)$ predicting noise at each timestep:

$$p_\theta(z_{t-1}|z_t) = \mathcal{N}(z_{t-1}; \mu_\theta(z_t, t), \Sigma_\theta(z_t, t)). \tag{4}$$

The model is trained with a simplified denoising loss [HJA20]:

$$L_{\text{denoise}} = \mathbb{E}_{z_0,\epsilon,t}\left[\|\epsilon - \epsilon_\theta(z_t, t)\|^2\right], \tag{5}$$

which minimizes the difference between predicted and true noise. This formulation enables high-quality generation but lacks explicit perceptual guidance, often leading to oversmoothed outputs.

*Adversarial Training in Latent Space.* Adversarial training has been widely explored in generative models, particularly in GANs [GPM*14]. Recent work on Energy-Based GANs (EB-GANs) [ZML17; KLL21] introduced an energy function that replaces the classifier-based discriminator, promoting structured sample separation. The discriminator assigns an energy value, where real samples have lower energy than generated ones, instead of classifying them as real or fake. The original EB-GAN was defined in pixel space, as discussed in Sec. 3.2. However, adversarial learning in pixel space may not fully leverage the structured representations in diffusion models. To overcome this limitation, we propose integrating adversarial training into the latent space of diffusion models.

*Connecting Denoising and Adversarial Learning.* Eq. (5) and the energy-based formulation in Sec. 3.2 provide the basis for our LD-GAN model. We introduce a novel training framework that enhances sample fidelity by combining likelihood-based training of diffusion models with structured adversarial objectives.

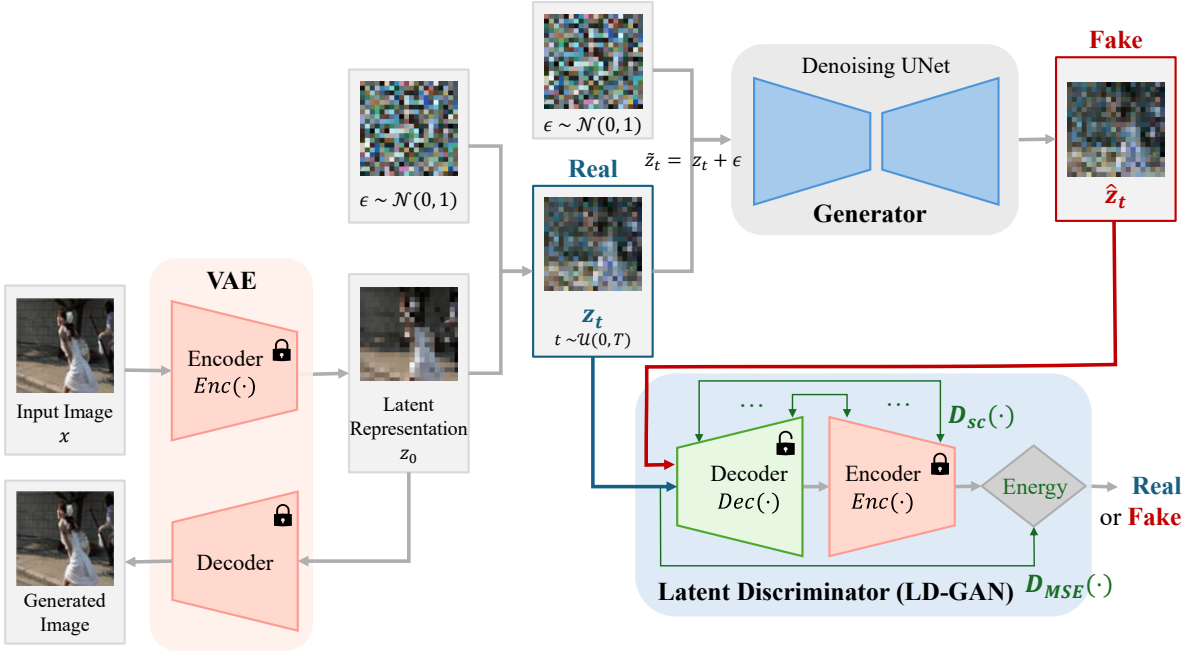

**Figure 1:** *Overview of LD-GAN. The generated latent $\hat{z}_t$ is produced by denoising $\tilde{z}_t = z_t + \epsilon$ using the generator (i.e., UNet). At each timestep $t$, both the real latent $z_t$ and the generated latent $\hat{z}_t$ are evaluated by an energy-based discriminator composed of VAE decoder and encoder. Unlike the standard encoder-decoder structure of VAE, the decoder learns noise-aware features while the frozen encoder anchors the latent code to the pretrained manifold.*

Although latent spaces support semantic abstraction and smooth interpolation [KW14; VKK21], enabling compact, diverse representations, they lack structural cues [DN21; ZIE*18]. The proposed method addresses this by defining energy over a decoder–encoder cycle. A latent code is decoded into a structure-aware representation and re-encoded to measure reconstruction consistency. This lets the discriminator capture both semantic and structural alignment, offering stronger inductive bias than latent-only methods. Our approach builds on energy functions in cycle-consistency models [DM19], motivated by latent EBMs [PHN*20] and cyclic energy models [XZL21], where structure-preserving reconstruction facilitates robust evaluation.

### 3.2 Adversarial Training with Latent Diffusion

In this section, we introduce *Latent Diffusion Generative Adversarial Networks (LD-GAN)*, a novel framework integrating adversarial training into latent diffusion models. LD-GAN differs from earlier adversarial learning methods using an additional discriminator network [WZH*23] by directly utilizing the variational autoencoder (VAE) in latent diffusion models as an *energy-based discriminator*. This enables adversarial training *without modifying the existing diffusion pipeline*, significantly reducing computational overhead.

*Overview of LD-GAN.* Standard diffusion models optimize an objective function based on likelihood, minimizing the denoising loss defined in (5). This framework allows high-quality image synthesis but lacks perceptual constraints, often leading to oversmoothed outputs. To address this, we introduce adversarial learning in the latent space, using an energy-based discriminator $D(z_t)$ to distinguish real and generated latent codes at randomly selected timesteps. Our method does not require a separate discriminator network, as we *integrate the existing VAE structure* from the latent diffusion model itself. The proposed method is illustrated in Fig. 1.

*Energy-Based Discriminator via VAE Integration.* Unlike conventional GANs that use a separate discriminator in pixel space, we define an energy-based discriminator by leveraging the decoder-encoder consistency in the VAE structure. Given a latent code $z$ at timestep $t$, which can be either the real latent $z_t$ or the generated latent $\hat{z}_t$, the discriminator energy is computed as follows:

$$D(z) = \frac{1}{d}\|z - \mathrm{Enc}(\mathrm{Dec}(z))\|^2, \tag{6}$$

where $d$ is the latent dimensionality, and $\mathrm{Enc}(\cdot)$ and $\mathrm{Dec}(\cdot)$ denote the VAE's encoder and decoder in the latent diffusion model.

Unlike the encoder–decoder structure of the standard VAE, our discriminator adopts a decoder-encoder structure. In this design, the discriminator receives a latent code $z_t$ at timestep $t$ as input, where $z_t$ inherently reflects the noise level introduced by the diffusion process. Since the noised latent $z_t$ follows a Gaussian distribution, it can be reliably represented through the pretrained VAE decoder and re-encoded by the encoder. This reversed flow allows the decoder in the discriminator to be fine-tuned to capture variations in noise levels, enabling the learning of timestep-aware features for $t > 0$ and improving its ability to provide informative adversarial feedback throughout the denoising process. In particular, the frozen encoder anchors the latent code to the pretrained latent manifold, providing a stable reference that guides decoder to learn noise-aware features across timesteps.

To further enhance the robustness to the noise level of $z_t$ in the timestep $t$, we replace group normalization with instance normalization, which normalizes per sample and better handles noise variation. Please refer Sec. 4.3.1 for more details. This normalization keeps energy invariant to latent dimensionality, ensuring stable training and consistent evaluation across timesteps. LD-GAN incorporates adversarial learning in the existing autoencoder framework *without modifying the original diffusion structure*.

*Efficient Training Without Interfering with Diffusion.* To prevent adversarial learning from disrupting the diffusion process, we *freeze the encoder* update only the decoder during training. This allows the model to leverage the structured latent space of the original VAE while improving sample realism via adversarial learning. The unchanged encoder preserves the generative prior of the original latent diffusion model, preventing overfitting to the training dataset. Our approach significantly reduces computational costs by introducing no extra trainable parameters beyond decoder fine-tuning.

*Training Objectives for LD-GAN.* The adversarial training in LD-GAN builds upon the EB-GAN framework [ZML17], adapting it to the latent space of diffusion models. We apply adversarial learning across randomly sampled timesteps $t \sim \mathcal{U}(0,T)$, ensuring that the model learns to distinguish real and generated latents across all diffusion stages. The discriminator loss is given by:

$$L_D = \mathbb{E}_{t\sim\mathcal{U}(0,T)}\left[D(z_t) + \max(0, m - D(\hat{z}_t))\right], \tag{7}$$

where $z_t$ represents the ground-truth latent representation at timestep $t$, $\hat{z}_t$ is the generator output at timestep $t$ predicted by the UNet of the latent diffusion model, and $m$ is a positive margin of the EB-GAN [ZML17].

*Generator Objective and Inference Stability.* The generator in LD-GAN follows a hybrid loss formulation, incorporating both the standard diffusion loss in (5) and an adversarial term:

$$L_G = L_{\text{denoise}} + \lambda_{\text{adv}}\mathbb{E}_{t\sim\mathcal{U}(0,T)}\left[L_{\text{adv},t}\right], \tag{8}$$

where $L_{\text{adv},t} = D(\hat{z}_t)$ encourages the generator to produce samples that receive lower energy from the discriminator at each randomly sampled timestep. The coefficient $\lambda_{\text{adv}}$ controls the balance between denoising and adversarial objectives. Furthermore, it is crucial to emphasize that *the decoder used for discriminator training i.e., fine-tuned decoder, is not used for image generation during inference*. During inference, we use the original VAE decoder of the latent diffusion model, ensuring that adversarial training does not affect the final image quality or the stability of the generation process. The theoretical proof of (7) and (8) is provided in Sec. 3.4.

*Stabilizing the Discriminator.* Since the discriminator (*i.e.*, the fine-tuned decoder and the fixed encoder) operates on latent representations at varying noise levels, it must adapt to different diffusion stages.

*Challenges in Training the Discriminator.* A unique challenge in training the discriminator is the variation in latent representations across different timesteps. Since adversarial learning is applied at randomly sampled timesteps, the feature distribution of $z_t$ changes dynamically, making it difficult for the discriminator to learn a stable decision boundary. To address this, we introduce two key stabilization techniques. First, we apply *instance normalization* [UVL16] within the discriminator (*i.e.*, fine-tuned decoder) to normalize feature statistics and reduce sensitivity to noise level variations. Unlike batch normalization, which operates across a batch of samples, instance normalization normalizes feature statistics at the individual sample level, preventing abrupt distribution shifts between different $z_t$. Second, we adopt a *multi-timestep training* strategy, where the discriminator is trained across randomly sampled timesteps $t \sim \mathcal{U}(0,T)$ rather than a fixed timestep. This prevents the discriminator from overfitting to specific noise levels and ensures that it generalizes across the entire diffusion process. These techniques significantly improve training stability and allow adversarial learning to remain compatible with the latent diffusion model.

*Discriminator Expressiveness and Integration with Diffusion.* Since LD-GAN's discriminator is derived from the pre-trained VAE, it benefits from a strong generative prior that enables better modeling of latent data distributions. Unlike conventional GAN discriminators, which learn a decision boundary from scratch, LD-GAN's discriminator leverages the structured latent space learned by the diffusion model. This results in improved sample quality and robust adversarial training.

### 3.3 Structural Consistency Energy

In adversarial training, the discriminator is crucial for distinguishing between real and generated latent representations. With a fixed encoder, the generator's ability to leverage the VAE prior depends on the alignment of encoder and decoder features. Furthermore, feature-based losses preserve perceptual quality by reinforcing structural attributes, preventing the excessive blurring common with pixel-based losses. Unlike traditional perceptual losses like

LPIPS [ZIE*18], which rely on a separately pretrained network, our structural consistency energy is defined within the autoencoder structure, requiring no additional model components. Unlike feature matching loss [SGZ*16], which relies on comparisons between real and generated samples, our method functions entirely within the generated latent space, making it independent of external reference data. To ensure structural consistency and effective integration with the pre-trained VAE, we introduce a *structural consistency energy* that extends the discriminator's evaluation criteria beyond pixel-wise reconstruction error.

*Defining Structural Consistency Energy.* The LD-GAN discriminator is an energy-based model that assesses latent consistency, as defined in (6). For convenience, we refer to this original reconstruction term as $D_{\text{MSE}}(z_t)$, ensuring pixel-wise consistency between the input latent and its reconstructed counterpart. This term regularizes reconstruction quality but does not ensure consistency across intermediate feature representations. To overcome this limitation, we introduce a structural consistency energy term $D_{\text{SC}}(z_t)$, defined as follows:

$$D_{\text{SC}}(z_t) = \frac{1}{|\mathcal{L}|} \sum_{l \in \mathcal{L}} \left( 1 - \frac{\left\langle f_{\text{Enc}}^{(l)}(z_t), f_{\text{Dec}}^{(l)}(\text{Dec}(z_t)) \right\rangle}{\|f_{\text{Enc}}^{(l)}(z_t)\| \cdot \|f_{\text{Dec}}^{(l)}(\text{Dec}(z_t))\|} \right), \quad (9)$$

where $\langle \cdot, \cdot \rangle$ denotes the dot product of two flattened feature vectors, and $f_{\text{Enc}}^{(l)}(\cdot)$ and $f_{\text{Dec}}^{(l)}(\cdot)$ represent the feature activations at the $l$-th layer of the encoder and decoder, respectively. Each $l$ corresponds to a matched resolution level in the encoder-decoder architecture, typically selected before encoder downsampling and after decoder upsampling. The set $\mathcal{L}$ contains layer pairs for alignment, ensuring the generator retains multi-scale structural information.

Our SCE term follows the feature alignment principle in perceptual and feature matching losses [SGZ*16; ZIE*18], where intermediate features are robust to low-level variations. Unlike external losses (*e.g.*, LPIPS [SGZ*16]), ours operates entirely within the autoencoder, enabling self-supervised regularization without distributional mismatch. The frozen encoder acts as a fixed perceptual backbone (*e.g.*, VGG in LPIPS), and the decoder is trained to align with the perceptual features. But, in our case, alignment is based on architectural symmetry: encoder features before downsampling are matched with decoder features after upsampling to ensure spatial and semantic consistency. Prior work has shown that such internal alignment improves training stability and representation quality [DM19; SGZ*16].

Thus, our structural consistency energy relies solely on the encoder-decoder structure of the latent diffusion model, requiring no external supervision. This approach is computationally efficient and improves perceptual fidelity.

As in prior perceptual and feature matching losses [SGZ*16; ZIE*18], feature layers are chosen empirically. A sensitivity analysis of feature layer configurations $\mathcal{L}$ for measuring structural consistency energy is presented in Appendix A.2.

*Integrating Structural Consistency into the Discriminator.* We redefine the total discriminator energy function with this feature-space constraint as follows:

$$D(z_t) = D_{\text{MSE}}(z_t) + \lambda_{\text{SC}} D_{\text{SC}}(z_t), \quad (10)$$

where $\lambda_{\text{SC}}$ is a parameter that controls the contribution of structural consistency. Incorporating feature-space constraints into the energy function allows the discriminator to evaluate samples with a richer set of perceptual criteria, improving adversarial robustness. Please refer to Sec.4.3.2 for an ablation study on structural consistency energy. A sensitivity analysis of the parameter $\lambda_{SC}$ is also in Sec. A.3.

*Impact on Generator Training and Sample Quality.* The generator optimizes the adversarially-extended denoising loss, formally defined in (8). The discriminator energy now incorporates structural consistency constraints from (9), prompting the generator to minimize both pixel-wise and feature-level discrepancies. This enhances perceptual quality by reducing blurriness and reinforcing edge structures, akin to LPIPS-based losses [ZIE*18]. However, unlike the previous perceptual losses like LPIPS that rely on a separately pretrained network, our method is self-contained in the diffusion model's autoencoder, reducing computational overhead. Since the encoder remains frozen, maintaining structural consistency allows the generator to retain compatibility with the structured latent space learned during VAE pretraining. As a result, adversarial updates remain integrated with the diffusion model's prior, preventing mode collapse and ensuring high-quality synthesis.

*Effect on Training Stability.* By explicitly incorporating structural consistency into the energy function, LD-GAN stabilizes the adversarial learning process and prevents the discriminator from relying solely on pixel-based errors. This structured sample evaluation approach reduces noise sensitivity and increases the consistency of learned latent representations. We further analyze the effect of structural consistency on perceptual quality in Sec. 4.3.2.

### 3.4 Theoretical Analysis of LD-GAN

In this section, we provide a formal proof that the proposed LD-GAN objective converges and enhances training stability and sample realism. Our proof leverages the convergence properties of Variational Autoencoders (VAEs) via ELBO optimization, the stability of Energy-Based GANs (EB-GANs), and the regularizing effect of the feature alignment loss.

*1. VAE Convergence via ELBO Optimization* Let $x$ be an input sample and $z$ its latent representation obtained from the encoder, *i.e.*, $z = \text{Enc}(x)$. The VAE is trained by maximizing the Evidence Lower Bound (ELBO):

$$\log p(x) \geq \mathbb{E}_{q(z|x)} [\log p(x|z)] - D_{\text{KL}}\Big(q(z|x) \,||\, p(z)\Big), \quad (11)$$

which guarantees that the learned latent distribution $q(z|x)$ converges towards the prior $p(z)$ under mild assumptions on the variational family. Thus, the VAE establishes a well-structured latent space that serves as a stable foundation for adversarial training. We emphasize that the convergence is local in nature, which is sufficient for stable training in typical deep learning practices.

*2. Convergence of the Denoising Loss* The training of the diffusion model involves minimizing the denoising loss.

$$L_{\text{denoise}} = \mathbb{E}_{z_0, \epsilon, t} \left[ \left|\left| \epsilon - \epsilon_\theta(z_t, t) \right|\right|^2 \right]. \quad (12)$$

Although the denoising loss is non-convex in practice, it has been empirically shown to converge under appropriate architectural choices and noise schedules [HJA20], supporting the assumption of local convergence in typical diffusion training settings. This empirical behavior aligns with practical findings in diffusion models, where convergence to high-quality solutions is frequently observed despite the underlying non-convexity. The (local) convexity of this loss ensures that the optimization process using standard gradient-based methods (*e.g.*, SGD) will reach a stationary point, provided that the learning rate and other hyperparameters are properly chosen.

*3. Stability via the EB-GAN Framework* LD-GAN adopts an energy-based discriminator:

$$D(z_t) = D_{\text{MSE}}(z_t) + \lambda_{\text{SC}} D_{\text{SC}}(z_t), \tag{13}$$

where $D_{\text{MSE}}(z_t)$ and $D_{\text{SC}}(z_t)$ are structured energy functions that ensure latent space consistency:

$$D_{\text{MSE}}(z_t) = \left\| z_t - \text{Enc}\left(\text{Dec}(z_t)\right) \right\|^2, \tag{14}$$

$$D_{\text{SC}}(z_t) = \sum_{l \in \mathcal{L}} \left\| f^{(l)}_{\text{Enc}}(z_t) - f^{(l)}_{\text{Dec}}\left(\text{Dec}(z_t)\right) \right\|^2. \tag{15}$$

As $D_{\text{MSE}}$ and $D_{\text{SC}}$ are defined as squared $\ell_2$-norm terms, the resulting energy $D(z_t)$ is guaranteed to be non-negative by construction, regardless of the specific choice of encoder and decoder. The EB-GAN formulation is known to be stable when its energy function is bounded and Lipschitz continuous [ZML17]. In practice, we rely on the mild assumption that the learned decoder and fixed encoder do not diverge significantly from the structured latent space. Since both $D_{\text{MSE}}$ and $D_{\text{SC}}$ are composed of smooth functions and squared norms of differentiable mappings (*i.e.*, encoder and decoder), their gradients are locally Lipschitz continuous under standard neural network assumptions [HRS16]. It ensures well-behaved gradient descent dynamics. As $D_{\text{MSE}}$ and $D_{\text{SC}}$ are squared-norm terms, they naturally remain non-negative and penalize large deviations.

*4. Regularization via Feature Alignment* Feature alignment loss further stabilizes adversarial training by enforcing consistency across hierarchical latent representations:

$$D_{\text{SC}}(z_t) = \sum_{l \in \mathcal{L}} \left\| f^{(l)}_{\text{Enc}}(z_t) - f^{(l)}_{\text{Dec}}\left(\text{Dec}(z_t)\right) \right\|^2. \tag{16}$$

This term prevents the decoder from deviating excessively from the pre-trained encoder features, effectively acting as a self-contained constraint that regularizes adversarial learning without requiring any external network [ZIE*18]. Consequently, multi-scale structures are preserved, reducing the risk of mode collapse and improving perceptual fidelity in generated samples.

*5. Combined Objective and Convergence Analysis* The overall generator objective is given by:

$$L_G = L_{\text{denoise}} + \lambda_{\text{adv}} D(z_t). \tag{17}$$

Since $L_{\text{denoise}}$ has guaranteed local convergence and $D(z_t)$ is non-negative and locally Lipschitz in the latent space (under the mild architectural assumptions above), we analyze the overall optimization dynamics. Specifically, it is observed:

- $L_{\text{denoise}} \geq 0$ is (locally) convex in $\epsilon_\theta$.
- $D(z_t) \geq 0$ is composed of squared-norm energy terms, thus lower-bounded by zero.
- The sum $L_G$ is therefore bounded below by 0.

These properties ensure that stochastic gradient descent (SGD) or its variants will converge to a stationary point. In other words, if $\{\theta_k\}$ is the sequence of generator parameters, then

$$\lim_{k \to \infty} \left\| \nabla L_G(\theta_k) \right\| = 0 \tag{18}$$

holds under standard conditions on the learning rate (*e.g.*, diminishing step size or sufficiently small constant step size for non-convex problems). Although the global optimum may not be guaranteed due to the non-convexity of deep networks, the local convergence is sufficient to ensure stable and high-quality synthesis in practice.

*6. Preservation of the Latent Space Structure* By freezing the encoder during adversarial training, the structured latent space learned via ELBO optimization remains intact. This avoids any distortion of the learned distribution during adversarial updates and ensures that the diffusion prior is preserved. Consequently, the decoder fine-tunes to improve sample realism but cannot alter the fundamental latent representation, fostering both stability and perceptual quality.

*Conclusion* By leveraging the convergence guarantees of VAE-based ELBO optimization, the regularization effects of feature alignment, and the stability of EB-GAN energy functions, we conclude that the LD-GAN objective:

$$L_G = L_{\text{denoise}} + \lambda_{\text{adv}} \left[ D_{\text{MSE}}(z_t) + \lambda_{\text{SC}} \sum_{l \in \mathcal{L}} \left\| f^{(l)}_{\text{Enc}}(z_t) - f^{(l)}_{\text{Dec}}\left(\text{Dec}(z_t)\right) \right\|^2 \right] \tag{19}$$

is composed of smooth, non-negative, lower-bounded components with known convergence behavior. This structure ensures that adversarial training in LD-GAN remains stable and effective while preserving the generative prior and improving perceptual quality. Thus, LD-GAN provides a theoretically grounded and empirically effective solution for stable adversarial training in latent diffusion models. □

## 4 Experiments

In this section, we conducted a series of experiments to evaluate the effectiveness and flexibility of the proposed method. In Sec. 4.2, we compared our approach with the baseline Latent Diffusion Model (LDM) [RBL*22], demonstrating the benefits introduced by our framework. Furthermore, we assessed the extensibility of our method across diverse diffusion-based tasks, including conditional text-to-image generation [LLW*23; GAA*22], and 2D-to-3D generation [LLZ*23; LLL*24], by comparing against state-of-the-art methods for each task. These results collectively highlight the adaptability and generalizability of our approach across a wide range of generative scenarios. Additionally, in Sec. 4.3, we conducted ablation studies to analyze the impact of instance normalization on the proposed energy-based discriminator and structural consistency energy in (9). Furthermore, for clarity, in Appendix A, we

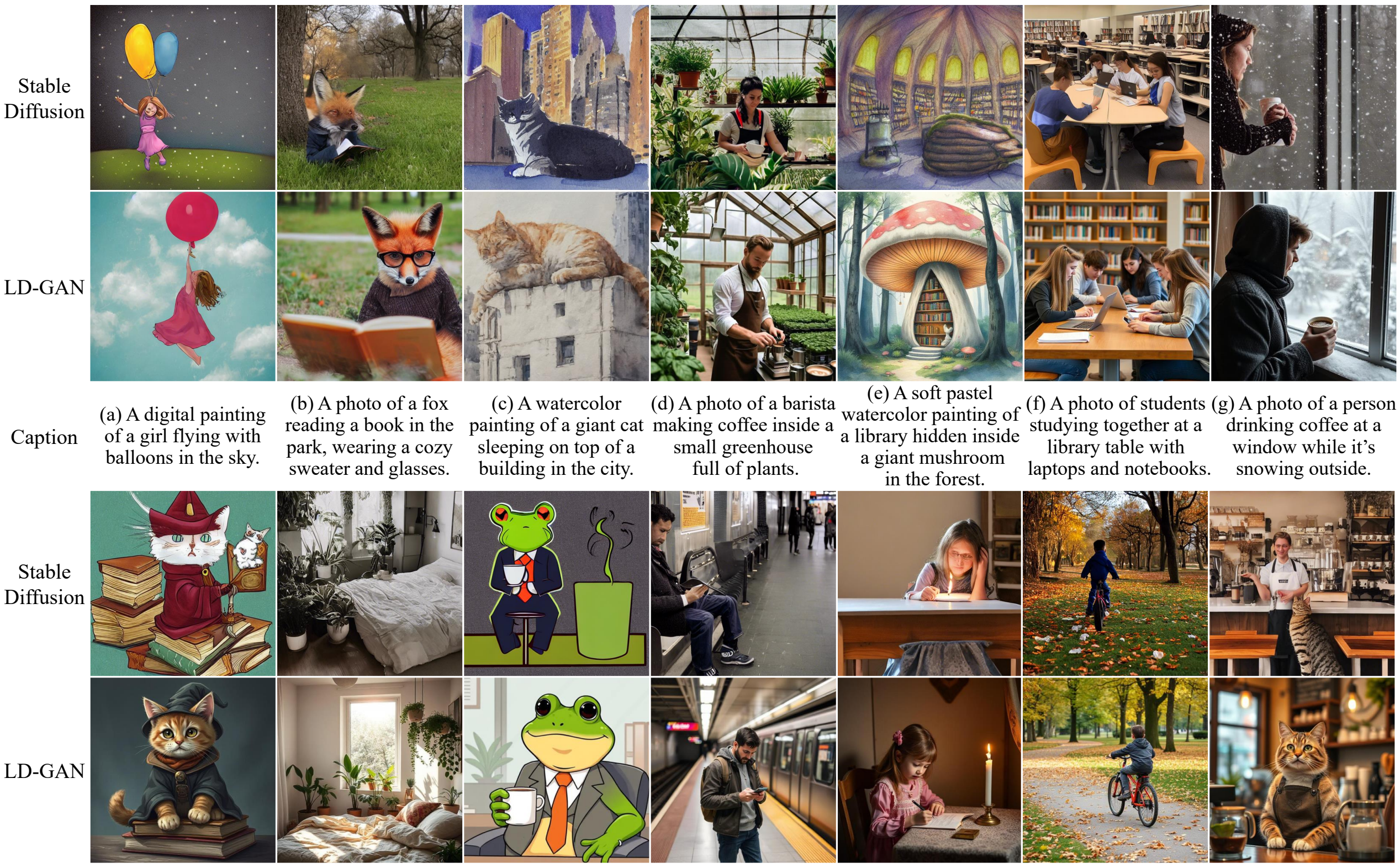

(a) A digital painting of a girl flying with balloons in the sky. (b) A photo of a fox reading a book in the park, wearing a cozy sweater and glasses. (c) A watercolor painting of a giant cat sleeping on top of a building in the city. (d) A photo of a barista making coffee inside a small greenhouse full of plants. (e) A soft pastel watercolor painting of a library hidden inside a giant mushroom in the forest. (f) A photo of students studying together at a library table with laptops and notebooks. (g) A photo of a person drinking coffee at a window while it's snowing outside.

(h) Fantasy art style of a cute cat wearing wizard clothes, sitting on books. (i) A photo of cozy bedroom interior, morning sunlight, plants, minimal design. (j) A comics of a frog wearing a business suit, holding a cup of coffee, sitting in an office. (k) A photo of a man waiting for the subway while checking his phone. (l) A photo of girl sitting at a small desk with a candle, writing a letter. (m) A boy riding a bicycle through a park full of fallen leaves. (n) A photo of a cat working as a barista in a cozy cafe, wearing an apron.

**Figure 2:** *Comparison with LD-GAN and Stable Diffusion [RBL*22] trained on the LAION-5B dataset [SBV*22].*

analyzed the sensitivity of the proposed method with respect to various hyperparameters, including the adversarial balancing parameter $\lambda_{adv}$ in (8), the structural consistency energy balancing parameter $\lambda_{SC}$ in (10), and the number of feature layers $|\mathcal{L}|$ in (9). These analyses further support the existence of a stable hyperparameter regime. Finally, we evaluated the convergence behavior and the training efficiency of the proposed LD-GAN in Appendix B. Additional qualitative analyses on the effect of the denoising objective and negative prompts are provided in Appendix C.

### 4.1 Experimental Details

All experiments, such as the baseline methods and the comparison experiments, were trained using the same number of steps and with the same number of parameters with 8 NVIDIA H100 GPUs. In particular, following baseline methods from recent works [RBL*22; LLW*23; GAA*22; LLZ*23; LLL*24] on conditional text-to-image and 2D-to-3D generations for fair comparisons, we employed the Stable Diffusion 2.1 (SD-2.1) [RBL*22] model as the base diffusion architecture for all experiments. Note that for computational efficiency, all experiments were conducted by fine-tuning the base diffusion architecture.

We set $\lambda_{adv} = 0.01$ in (8) and $\lambda_{SC} = 0.05$ in (10) for all experiments, showing the best performance. A sensitivity analysis of the adversarial parameter $\lambda_{adv}$ is provided in Appendix A.1, and the structural consistency energy parameter $\lambda_{SC}$ is analyzed in Appendix A.3. Also, we select the set $|\mathcal{L}| = 3$ in (9) for all experiments, which achieved the best empirical result. Here, we selected three layers right before downsampling in the encoder and the three layers immediately after upsampling in the decoder as corresponding pairs. A sensitivity analysis of feature layers $|\mathcal{L}|$ used for measuring the structural consistency energy is presented in Appendix A.2 of the *supplemental material*.

Following previous works on energy-aware adversarial optimization [ZML17; GWJ*20], the margin parameter $m$ in the discriminator loss defined in (7) was determined by periodically computing the average reconstruction error on the training set during inference, specifically every 5 epochs in our implementation. This approach is grounded in the decision-theoretic principle that the conditional mean is the optimal estimator under a squared error loss function [LC98]. By setting $m$ to the empirical mean of the reconstruction errors, the margin is adaptively aligned with the data distribution, avoiding both overestimation and underestimation. This ensures a consistent separation between real and generated latent representations, enabling the discriminator to maintain a stable decision boundary throughout training. In our tests, perturbing the margin value within a ±10% range did not lead to noticeable performance differences, suggesting that the proposed method remains robust under moderate variations of $m$.

**Table 1:** *Evaluation of performance comparisons with SD-2.1 [RBL*22] trained on the LAION-5B dataset [SBV*22].*

| Method | FID (↓) | CLIP score (↑) |
|---|---|---|
| SD-2.1 | 13.79 | 0.3098 |
| + LD-GAN (Ours) | **9.42** | **0.3428** |

## 4.2 Performance Comparisons

### 4.2.1 Text-to-Image Generation.

To evaluate the performance of the proposed method, we conducted experiments in comparison with the baseline model, Stable Diffusion [RBL*22]. For a fair comparison, all models were trained on the same LAION-5B dataset [SBV*22]. For evaluation, we randomly sampled 20K image-text pairs from the COCO2014 validation set [LMB*14] and the Flickr30k dataset [YLHH14]. Following prior works [ZRA23; BNH*22], we adopted two widely used metrics: the Fréchet Inception Distance (FID) [HRU*17], assessing distributional similarity between real and generated images, and the CLIP score [RKH*21], evaluating semantic alignment between text prompts and generated images through normalized CLIP embeddings.

**Results with SD-2.1.** First, we compared our method with SD-2.1 [RBL*22], which is used in other generation scenarios, such as conditional text-to-image generation in Sec. 4.2.2 and 2D-to-3D generation in Sec. 4.2.3. Both SD-2.1 and LD-GAN were trained for 2M steps over 17 days, using a batch size of 1024 and a learning rate of $1 \times 10^{-4}$.

Table 1 presents comparative performance results against standard Stable Diffusion [RBL*22], evaluated in terms of the FID and the CLIP score. The results demonstrate that incorporating the proposed method into Stable Diffusion consistently improves both distributional similarity and text-image alignment. Specifically, the improvement in CLIP score confirms that the proposed method enhances semantic consistency with input prompts. This indicates that adversarial training enables the method to better capture the underlying data distribution, leading to higher fidelity and semantically aligned generations. These improvements are evident in scenarios with complex visual concepts or fine-grained textual prompts, where the baseline often produces over-smoothed or ambiguous outputs. In contrast, the proposed method generates sharper details and more semantically faithful content, demonstrating the effectiveness of latent adversarial learning.

Qualitative comparisons further supporting this trend are illustrated in Fig. 2. The results show that the proposed LD-GAN demonstrates strong performance in generating plausible and high-quality images for imaginative or uncommon prompts. For instance, in Fig. 2 (b), our method synthesizes a coherent, detailed scene from the prompt *"a fox reading a book in the park, wearing a cozy sweater and glasses."*. In contrast, the baseline fails to represent the correct object composition and omits fine details, such as glasses and fabric texture, while our method accurately captures all elements, resulting in a high-frequency image with rich semantic detail. Also, in Fig. 2 (n), the scene of "a cat working as a barista in a cozy cafe" is depicted with higher fidelity and fewer visual artifacts compared to the baseline. These results suggest that LD-GAN can generalize effectively to visually novel and abstract scenarios. The advantages of our approach are evident in complex prompts. In Fig. 2 (c), our method generates a well-structured watercolor painting of a *"giant cat sleeping on top of a building"*, faithfully reflecting the caption while maintaining artistic fidelity. Similarly, in Fig. 2 (f), describing a multi-person scene, our method maintains anatomical correctness, facial expressions, and realistic object interactions, outperforming the baseline, which produces distorted or oversimplified compositions. Moreover, the proposed method demonstrates a significant advantage in accurately reconstructing fine-grained semantic elements and human structures. For instance, in Fig. 8 (k), the individual waiting for the subway is depicted with more natural body postures and facial alignments compared to the baseline. Similarly, in Fig. 8 (l), the girl's pose, candle placement, and expression appear more contextually coherent and detailed.

**Table 2:** *Evaluation of performance comparisons with SD-3.5-M [PEL*24] trained on the LAION-5B dataset [SBV*22].*

| Method | FID (↓) | CLIP score (↑) |
|---|---|---|
| SD-3.5-M | 7.983 | 0.3521 |
| + LD-GAN (Ours) | **6.732** | **0.3705** |

**Results with SD-3.5.** Although we employed SD-2.1 model to align with recent works [RBL*22; LLW*23; GAA*22; LLZ*23; LLL*24], such as conditional text-to-image and 2D-to-3D generation, we further apply our method to the state-of-the-art text-to-image generation model [PEL*24]. We employed the state-of-the-art Latent Diffusion model, *i.e.*, Stable Diffusion 3.5 Medium (SD-3.5-M) [PEL*24] as the baseline.

To adapt the proposed method to the SD-3.5 architecture, we first modified the discriminator (*i.e.*, the VAE decoder) to match the 16-channel latent representation used by SD-3.5, expanding the original 4-channel configuration to 16 channels. We also incorporated the VAE scaling factor (1.5305) into both the encoding and decoding processes to ensure consistency with the pretrained latent space. The generator was kept identical to the original SD-3.5 denoiser (*i.e.*, MMDiT-X) without architectural changes. To mitigate sudden gradient fluctuations caused by the increase in the number of parameters, we reduced adversarial loss weight to $\lambda_{\text{adv}} = 0.005$ and the learning rate to $5 \times 10^{-5}$. The SCE weight was kept at $\lambda_{\text{SCE}} = 0.05$, consistent with the main experiments, to maintain structural alignment across scales. Both SD-3.5-M and LD-GAN were trained for 2M steps over 22 days, using a batch size of 1024 and a learning rate of $1 \times 10^{-4}$.

Table 2 compares the performance of SD-3.5-M with and without the proposed method. The results show that the integration of LD-GAN into SD-3.5-M produces substantial improvements in all evaluation metrics, aligned with the trends observed in the SD-2.1 case in Table 1. This consistent behavior confirms that the proposed method provides robust performance gains even when applied to the latest state-of-the-art models, underscoring its strong generalization capability and effectiveness regardless of the underlying architecture or model design.

**Table 3:** *Evaluation of comparisons with conditional text-to-image methods [LLW*23; GAA*22].*

| Method | FID (↓) | CLIP score (↑) | CLIP similarity (↑) |
|---|---|---|---|
| GLIGEN | 11.97 | 0.286 | - |
| + LD-GAN (Ours) | **9.03** | **0.318** | - |
| Textual Inversion | 12.65 | - | 0.754 |
| + LD-GAN (Ours) | **8.43** | - | **0.792** |

#### 4.2.2 Conditional Text-to-Image Generation.

To validate the flexibility and extensibility of the proposed method, we evaluated LD-GAN on two state-of-the-art conditional text-to-image generation frameworks: GLIGEN [LLW*23] and Textual Inversion [GAA*22], comparing performance with and without our method. GLIGEN and Textual Inversion represent distinct forms of conditional generation: the former incorporates explicit spatial conditions such as bounding boxes, while the latter enables conditioning on novel visual concepts learned from just a few example images.

We employed FID [HRU*17] to evaluate distributional similarity in GLIGEN and Textual Inversion experiments. For GLIGEN, we additionally used the CLIP score [RKH*21] to measure semantic alignment between generated images and their textual descriptions. In the case of Textual Inversion, we adopted CLIP similarity [GPM*22], defined as the average pairwise cosine similarity between the CLIP image embeddings of input and generated images, to assess visual consistency in few-shot conditions.

**Results with GLIGEN.** GLIGEN extends traditional text-to-image diffusion models by incorporating additional grounding inputs, such as bounding boxes, keypoints, and semantic maps. We evaluated LD-GAN's impact on mode coverage by experimenting with GLIGEN conditioned on bounding boxes and captions. Both variants were trained on Object365 [SLZ*19] and GoldG [LZZ*22a]. For evaluation, we sampled 2K image-text pairs from the Flickr30k [YLHH14] and SBU [OKB11] datasets, with bounding boxes obtained using GLIP [LZZ*22b]. Both GLIGEN and LD-GAN were trained for 400k steps over 18 hours, using a batch size of 64 and a learning rate of $5 \times 10^{-5}$.

Table 3 presents the performance comparison of GLIGEN with and without the proposed method, using FID and the CLIP score. The results show that GLIGEN integrated with LD-GAN consistently outperforms the baseline across all evaluation metrics. In particular, lower FID indicates improved image fidelity and better alignment with the underlying data distribution, while the higher CLIP score suggests stronger semantic consistency between generated images and text prompts. Interestingly, LD-GAN enhances both visual realism and semantic grounding without additional conditions or supervision. This shows that adversarial training in latent space generates plausible images that align more closely with multimodal inputs. Moreover, the increase in the CLIP score indicates that the adversarial objective optimizes structural cues (*e.g.*, bounding boxes) and enhances holistic conditioning on text and spatial inputs. This balanced conditioning leads to better mode coverage and lowers the risk of overfitting to specific modalities.

These trends are further supported by the qualitative comparisons in Fig. 3, where LD-GAN yields coherent object placement and sharper visual attributes while maintaining semantic relevance to the captions and bounding box annotations. For example, in the first example, where the caption is *"An astronaut is reading a book while sitting on a giant hamburger floating in the ocean next to a seagull,"* the baseline method struggles to harmonize object positioning. The elements appear crowded and incoherent, suggesting that the model has overfit to bounding box locations without adequately capturing semantic relationships. In contrast, the proposed method produces a balanced, semantically accurate image, with the astronaut, hamburger, and seagull naturally placed and caption details clearly reflected, including fine textures like the book and ocean background. Similar advantages arise in the second and third examples. In the second case, our method accurately places *"a tree"* next to *"a bench"* and *"a man,"* while the baseline misrepresents spatial relationships, leading to visual inconsistencies. In the third sample, the red post office box is accurately reconstructed in the proposed method, including fine structural features and appropriate color distribution, while the baseline output lacks sharpness and contextual integration. Additionally, in the fourth example, the prompt describes a surreal scene: "A wizard sitting on a donut floating in a lake with a glowing wand as ducks swim by." While GLIGEN struggles to generate key elements, producing the wizard with an incoherent layout, our proposed method faithfully renders all elements, "wizard," "donut," "lake," "glowing wand," and "ducks", with semantically appropriate placement and sharp structural consistency. This observation indicates that LD-GAN effectively integrates complex scene elements while maintaining spatial realism. Furthermore, in the fifth case, the GLIGEN output exhibits anatomical distortions and oversmoothing, particularly in the facial features and overall structure of the human figure. In contrast, LD-GAN produces a coherent human figure with clear facial expressions and structurally accurate rendering of the "scooter" and "traffic light," preserving both detail and spatial layout.

**Results with Textual Inversion.** Textual Inversion enables the learning of new visual concepts in the text embedding space of diffusion models with minimal input images. We evaluated LD-GAN's impact on generation quality and stability with limited data through comparative experiments. The Google Scanned Objects (GSO) dataset and a custom multi-view dataset were used, selecting seven 512×512 images per concept. Both Textual Inversion and LD-GAN, were trained for 5k steps over 2.1 hours, using a batch size of 4 and a learning rate of $5 \times 10^{-3}$.

Table 3 presents the performance of Textual Inversion, with and without the proposed method, evaluated using FID and CLIP similarity. The results demonstrate that the incorporation of LD-GAN leads to consistent improvements in both metrics. In particular, the substantial reduction in FID highlights that adversarial training enables the method to better capture the data distribution, even under limited supervision, resulting in more realistic and high-fidelity samples. Furthermore, the observed improvements in CLIP simi-larity indicate that LD-GAN effectively preserves semantic identity and visual consistency of the input concept, which is critical in few-shot generation settings, such as Textual Inversion. This suggests that the proposed method improves both generalization and stability during training, even when the available data is minimal.

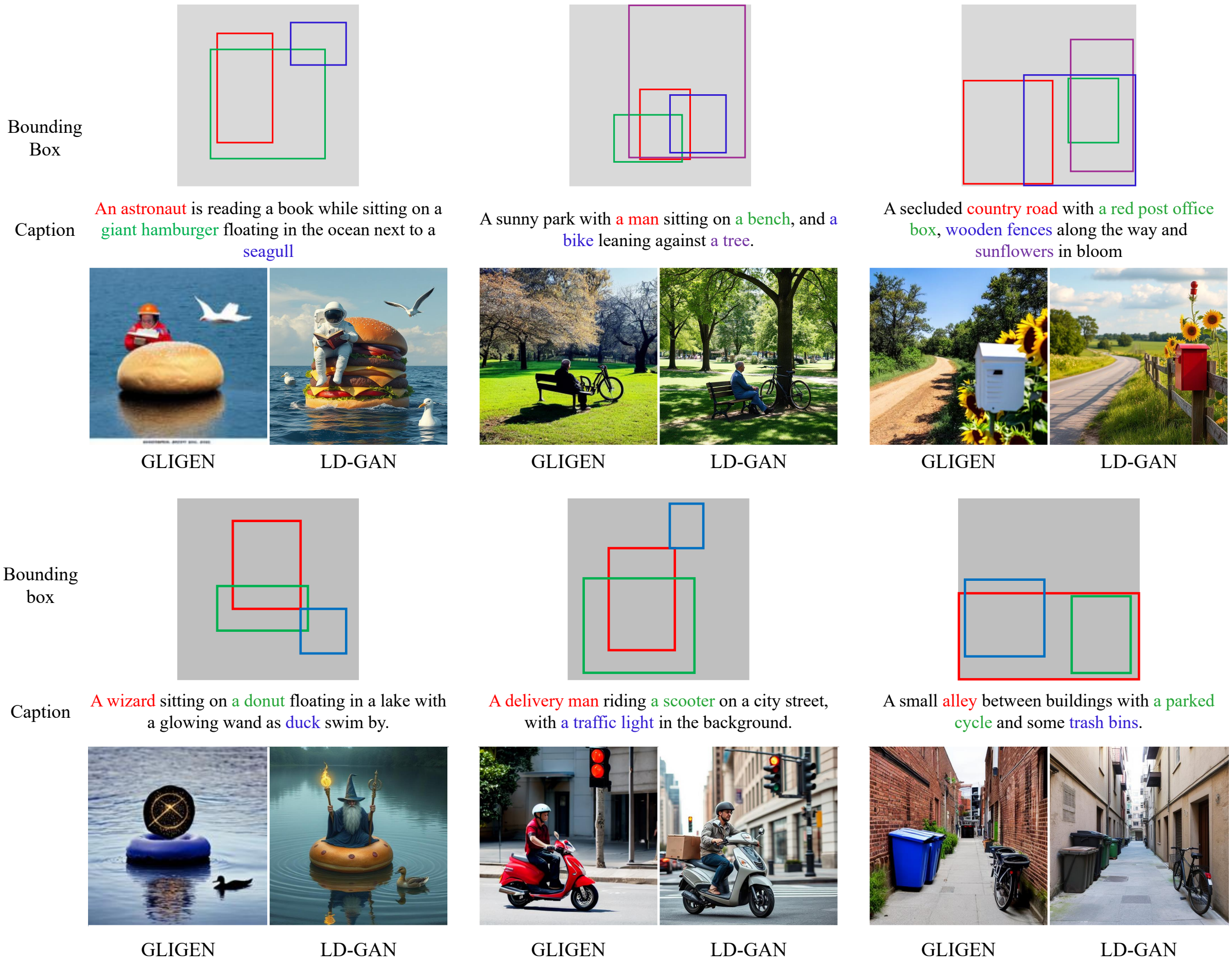

**Figure 3:** *Sample comparisons with the proposed method and GLIGEN [LLW*23] for image generation conditioned on bounding boxes. Both methods are trained on the Object365 dataset [WZF*23] and the GoldG dataset [LZZ*22a].*

**Table 4:** *Performance comparisons with multi-view images and textured mesh generated by 2D-to-3D methods [LLZ*23; LLL*24].*

| **Metho** | **PSNR (↑)** | **SSIM (↑)** | **LPIPS (↓)** | **CD (↓)** | **IoU (↑)** |
|---|---|---|---|---|---|
| SyncDreamer | 20.17 | 0.756 | 0.136 | 0.0232 | 0.5354 |
| + LD-GAN (Ours) | 22.32 | 0.846 | 0.115 | 0.0154 | 0.6749 |
| Era3D | 21.05 | 0.789 | 0.138 | 0.0174 | 0.6513 |
| + LD-GAN (Ours) | **23.83** | **0.837** | **0.116** | **0.0125** | **0.7231** |

Figure 4 provides visual comparisons, showing that LD-GAN contributes to sharper details, improved structure preservation, and semantic alignment with input images. For instance, in the first example of Fig. 4, the baseline method shows noticeable limitations when adapting the learned concept to various styles, such as *"the beach"* or *"adventure fantasy style."* It tends to overfit the input sample's geometric structure, resulting in outputs that fail to harmonize with the prompts. In contrast, the proposed method effectively integrates the input image with desired stylistic attributes, producing sharper and semantically aligned results across all variations. The advantage of LD-GAN is further illustrated in the second sample. While the baseline produces inconsistent geometry across different prompts, such as distorted posture and facial shape inconsistencies, our method reliably maintains the input's structural integrity, even with varied contexts like "a paper with a $\mathcal{S}_*$ shaped stamp mark" or "a small golden $\mathcal{S}_*$ on a wooden desk.". Moreover, as shown in the fourth sample, the baseline tends to overfit to the appearance of the input samples, resulting in artifacts where patterns bleed unnaturally across object boundaries. For instance, the subject's face becomes patterned under the prompt "a teacup in the style of $S_*$." In contrast, the proposed method maintains structural and perceptual separation between the concept and style, leading to natural, visually pleasing outputs that accurately reflect both the identity of the concept and the intended artistic transformation. These results suggest that LD-GAN effectively learns a robust concept representation that generalizes well across varying compositions and artistic domains.

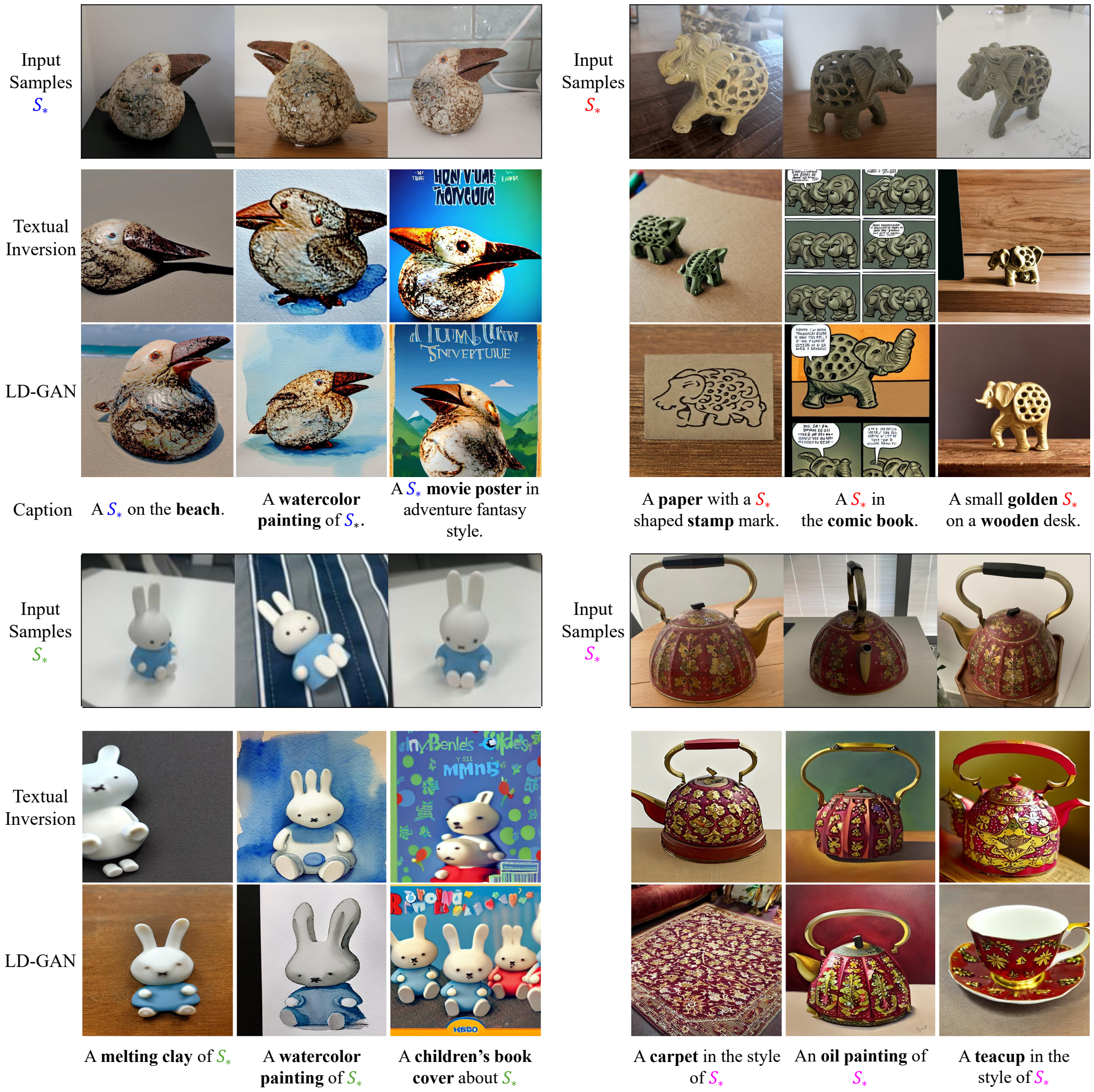

**Figure 4:** *Comparisons with Textual Inversion [GAA*22]. We trained with 7 input images per object provided by the user. $S_*$ is a pseudo-word representing the embedding vector for the input concept.*

#### 4.2.3 2D-to-3D Generation.

To evaluate the *extensibility* of the proposed method, we conducted comparative experiments using recent state-of-the-art 2D-to-3D generation models [LLZ*23; LLL*24] with and without LD-GAN. Specifically, we adopted SyncDreamer [LLZ*23] and Era3D [LLL*24] as baseline models and assessed their performance when integrated with our proposed adversarial training framework.

We evaluated both *novel view synthesis* and *3D reconstruction* to assess the generative performance of the models. For performance measurements of novel view synthesis, we used PSNR, Structural Similarity Index Measure (SSIM) [WBSS04], and LPIPS [ZIE*18]. For 3D reconstruction, we followed recent works [LWV*23; LGL*24] and used Chamfer Distance (CD) and Volume Intersection over Union (IoU) as evaluation metrics.

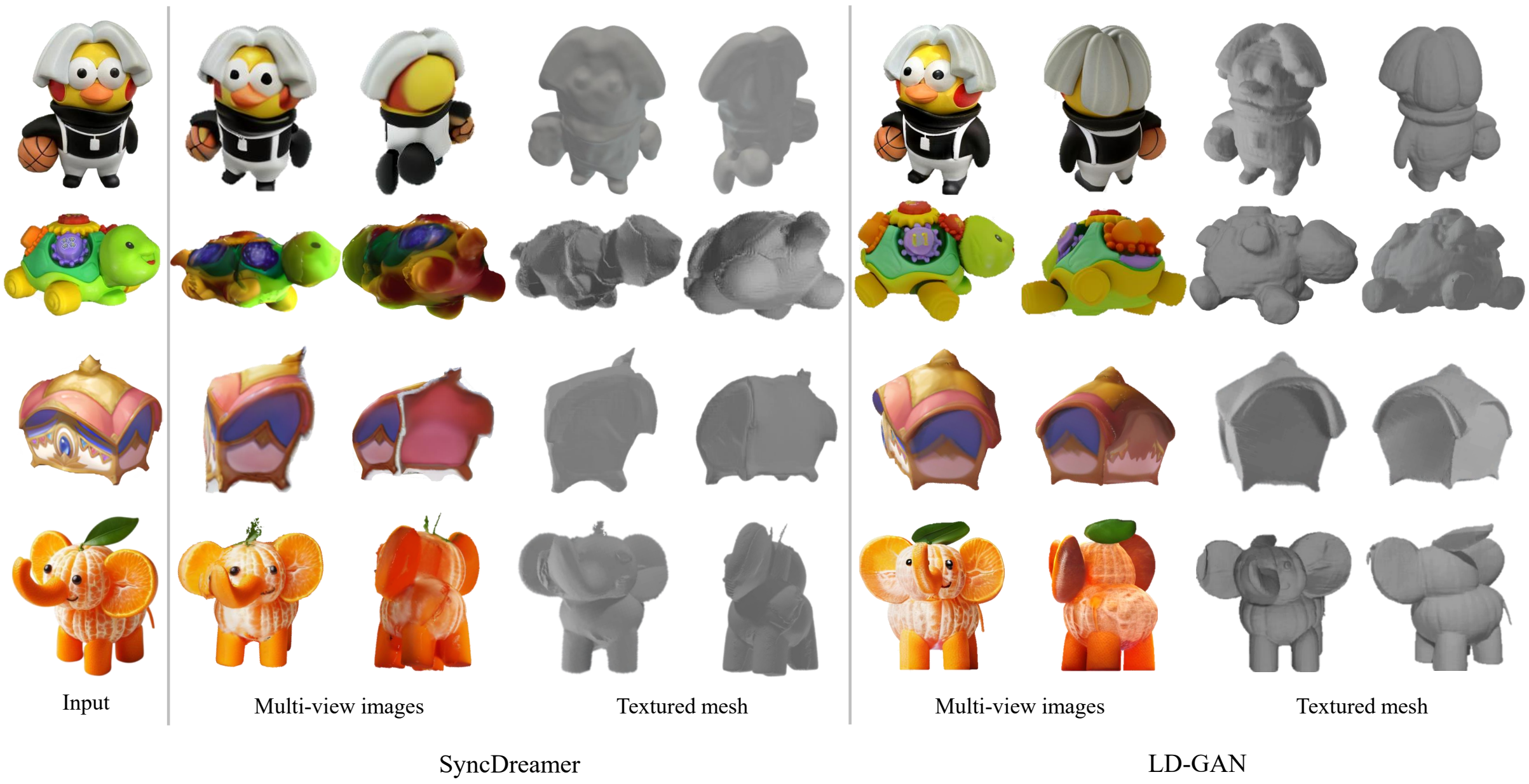

**Figure 5:** *Comparisons of multi-view images and textured mesh generated with SyncDreamer [LLZ*23].*

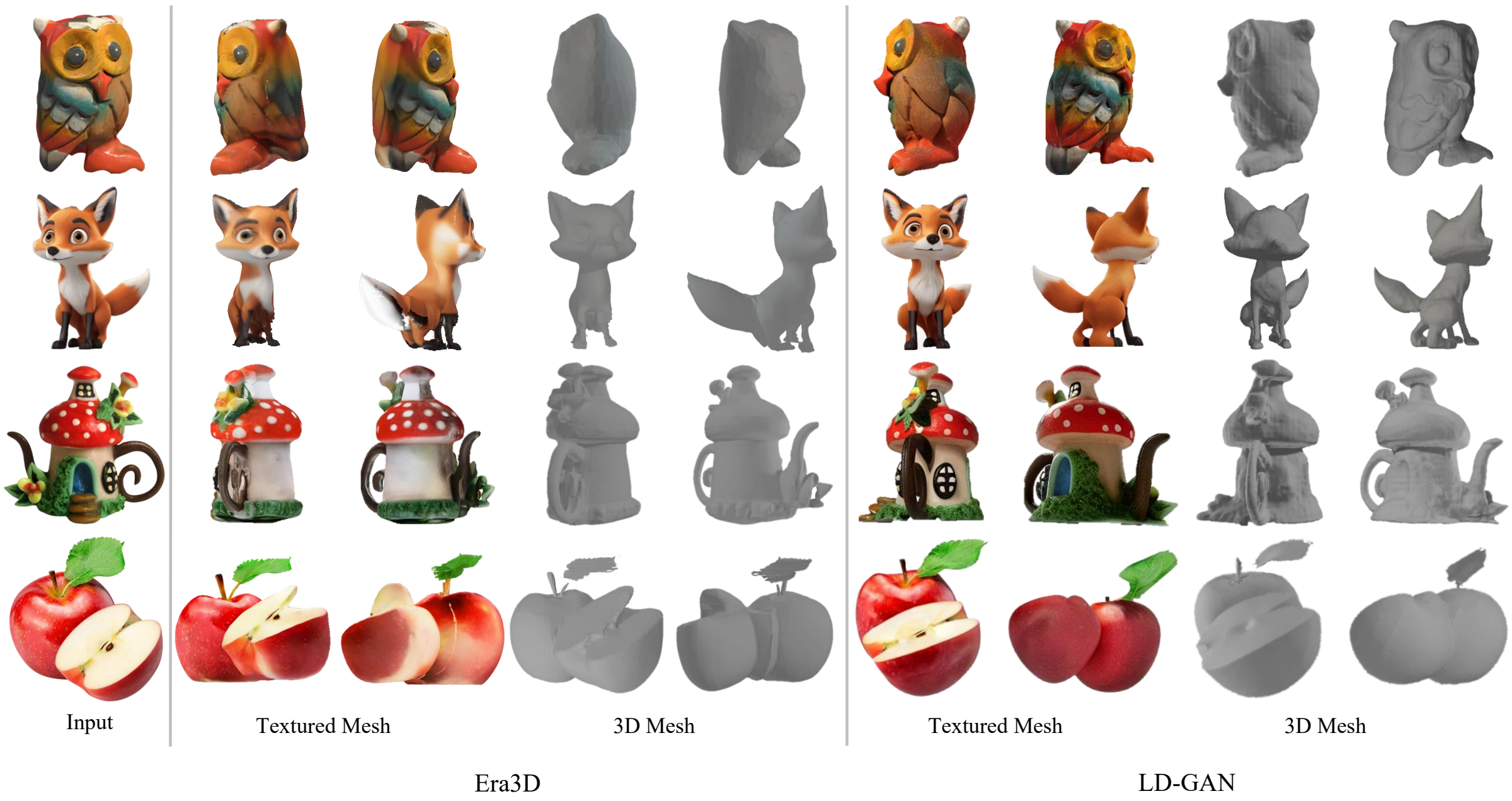

**Figure 6:** *Comparisons of multi-view images and textured mesh generated with Era3D [LLL*24].*

**Results with SyncDreamer.** SyncDreamer synthesizes consistent multi-view images from a single-view input using a 3D-aware feature attention mechanism during diffusion, enabling synchronized states across views in reverse sampling. In contrast, Era3D predicts camera parameters and employs row-wise attention to impose epipolar priors for multi-view image generation from a single input view. All models were trained using renderings from a subset of the Objaverse dataset [DSS*23], with each image at 512×512 resolution. To evaluate SyncDreamer, with and without LD-GAN, we used 32 training views per object: 16 images with uniform azimuths and fixed elevation of 30°, and 16 images with uniformly sampled azimuths and randomly sampled elevations from the range $[-20^\circ, 40^\circ]$. Both SyncDreamer and LD-GAN were trained for 80k steps over 39 hours, using a batch size of 128 and a learning rate of $5 \times 10^{-4}$ to $1 \times 10^{-5}$.

Table 4 summarizes the performance of SyncDreamer with and without the proposed method. The results indicate that the proposed method outperforms the baseline in both *novel view synthesis* and *3D reconstruction* tasks. In particular, the notable improvements in PSNR, SSIM, and LPIPS for the novel view synthesis task indicate that adversarial training guided by the structural consistency energy enables the decoder to more effectively capture geometric and semantic consistency across multi-view images. This results in sharper, more coherent, and view-consistent generations. Moreover, for 3D reconstruction, improvements in CD and IoU demonstrate that the proposed method facilitates better 3D structure reasoning from limited 2D observations. This improvement is credited to the discriminator's ability to enforce multi-scale feature alignment during training, which encourages the generator to learn spatial priors more effectively. These findings suggest that LD-GAN not only enhances visual fidelity but also improves the geometric plausibility of generated content.

The qualitative comparison results in Fig. 5 confirm that LD-GAN improves shape preservation and texture continuity across viewpoints, particularly in challenging poses or occlusions. For example, in the first row, the proposed method generates more consistent multi-view images, especially in rear views preserving fine details, such as clothing folds and hair shapes. In contrast, the baseline method often lacks consistency, resulting in blurred or distorted reconstructions. This advantage extends to the final textured mesh, where our method achieves more refined and volumetrically accurate outputs. The second row highlights another key difference in lighting and color consistency. While SyncDreamer inaccurately captures lighting characteristics, causing color shifts and artifacts in multi-view images and textured meshes, our method effectively preserves appearance across views, resulting in smoother transitions and more photorealistic reconstructions. Furthermore, in the fourth row of Fig. 5, the object exhibits complex surface textures and occlusions (*e.g.*, an orange with leaves), which results in a loss of texture fidelity. Conversely, our method effectively captures high-frequency visual features, such as the rough surface of the orange and the leaf structure, across all views. These improvements are clearly reflected in the corresponding textured mesh, where our reconstruction presents a significantly more faithful rendering with realistic surface topology and detailed texture alignment.

**Results with Era3D.** For Era3D, training was conducted with orthographic and perspective views. Specifically, 16 images were rendered using an orthographic camera with uniformly sampled azimuths and a fixed elevation of $0^\circ$. For each azimuth, three perspective views with elevations $[-20^\circ, 40^\circ]$ and one orthographic view were randomly sampled for training. Both Era3D and LD-GAN were trained for 40k steps over 53 hours, using a batch size of 128 and a learning rate of $1 \times 10^{-4}$ to $5 \times 10^{-5}$. For evaluation, we used single-view renderings from GSO [DFK*22] and OmniObject3D (Omni3D) [WZF*23].

Table 4 compares the performance of Era3D with and without the proposed method. The results demonstrate that integrating LD-GAN into Era3D leads to a substantial improvement across all evaluation metrics. In particular, the observed gains in CD and IoU highlight significant enhancement in 3D reconstruction quality. These improvements suggest that the proposed method, through structural consistency energy, encourages the generation of multi-view images that preserve geometric integrity and structural alignment, which are critical for accurate 3D shape recovery. Furthermore, the adversarial training with feature-space constraints promotes sharper and semantically coherent views, enhancing the resolution of ambiguous geometries and depth perception.

Figure 6 demonstrates that LD-GAN yields more realistic and structurally consistent reconstructions, particularly in challenging viewpoints and object shapes. In particular, in the first row of Fig. 6, the proposed method captures localized features, such as feathers and claws, more accurately than Era3D, which suffers from oversmoothing and loss of semantic clarity in multi-view images and textured meshes. This indicates that LD-GAN provides more expressive feedback through latent-space adversarial learning, leading to perceptually richer generation. Similarly, the second row highlights our method's effectiveness in preserving textures and structures, such as chest fur and tail coloration. In contrast, Era3D struggles with view consistency, resulting in texture inconsistencies and structural deformation. These differences are evident in the reconstructed meshes, with LD-GAN producing sharper and more geometrically accurate outputs, whereas Era3D insufficiently recovers intricate surface details. Moreover, our method effectively preserves global structure and appearance-related information, such as color and lighting, across all viewpoints. For instance, in the third row, the baseline method fails to reconstruct the correct geometry of the stairs, particularly from the side view, and generates inconsistent lighting conditions across views. In contrast, our method produces more coherent and accurate multi-view images, exhibiting consistent color tones and shading, which results in a more faithful textured mesh. These results collectively validate the proposed method's extensibility to complex 2D-to-3D generation tasks and its ability to enhance both perceptual realism and geometric fidelity in multi-view generative models.

## 4.3 Ablation Studies

### 4.3.1 Analysis on Instance Normalization

In this section, we analyzed the impact of instance normalization [UVL16] on the performance and stability of the proposed energy-based discriminator. To this end, we conducted an ablation study comparing models trained with and without instance normalization in the decoder of the variational autoencoder (VAE) used as the discriminator. Importantly, the original VAE encoder and decoder architecture of Stable Diffusion [RBL*22] adopted group normalization [WH18] by default. In our implementation, we replaced group normalization with instance normalization in the *decoder*, while keeping the encoder frozen throughout training. This modification is intended to stabilize adversarial learning across timesteps during latent diffusion. To ensure a fair comparison, we trained all variants using a fixed subset of the LAION-5B dataset [SBV*22], and kept all other hyperparameters and experimental settings identical to those described in Sec. 4.1. For evaluation, we randomly sampled 20K image-text pairs from the COCO2014 validation set [LMB*14] and the Flickr30k dataset [YLHH14], following the same proto-col as in Sec. 4.2.1. Generation quality was assessed using the FID [HRU*17] and the CLIP score [RKH*21].

**Table 5:** *Ablation study on the effect of normalization in the VAE decoder used as the discriminator.*

| Method | FID (↓) | CLIP score (↑) |
|---|---|---|
| Group normalization | 11.02 | 0.3183 |
| Instance normalization (Ours) | **9.42** | **0.3428** |

The comparison results are summarized in Table 5. We observe that using instance normalization in the decoder leads to clear improvements in both metrics. These results indicate that the introduction of instance normalization leads to more stable and effective adversarial training. This improvement can be attributed to the characteristics of the diffusion process. During training, the energy-based discriminator evaluates latent features $z_t$ sampled at random timesteps. Without instance normalization, the feature distributions across different $z_t$ may vary significantly due to the varying noise levels. Such distributional shifts hinder the discriminator's ability to model a consistent latent space, leading to unstable adversarial gradients, oscillatory generator updates, and even mode collapse.

In contrast, instance normalization normalizes feature statistics on a per-sample basis. This per-instance normalization helps reduce the effect of timestep-dependent variance in $z_t$, enabling the discriminator to operate over a smoother and more coherent latent space. As a result, the adversarial loss becomes more stable, leading to improved generator-discriminator interactions and enhanced training robustness across diverse noise levels and sample distributions. Furthermore, although instance normalization is only applied to the decoder, it significantly improves the effectiveness of the feature alignment process. By stabilizing the decoder-side representations, instance normalization facilitates more accurate multi-scale feature alignment with the fixed encoder. This enables better preservation of structural attributes and supports the learning of perceptually coherent outputs. Consequently, the generated samples exhibit higher sharpness and greater structural consistency.

We also empirically observed that these benefits generalize across diverse generative tasks, such as conditional text-to-image generation, and 2D-to-3D generation. While detailed quantitative results are provided for the text-to-image setting, similar qualitative improvements were consistently observed in other domains, indicating that instance normalization contributes to enhanced structural coherence and perceptual fidelity across various tasks. These findings highlight the practical value of replacing group normalization with instance normalization in the context of latent adversarial learning. In particular, our approach demonstrates that even applying instance normalization solely in the decoder without modifying the encoder can yield substantial benefits by stabilizing the discriminator and enhancing the generator's ability to learn complex structural patterns from noisy latent inputs.

### 4.3.2 Structural Consistency Energy Analysis

In this section, we analyze the impact of the structural consistency energy defined in (9) on the performance of the proposed method by comparing it with a conventional perceptual loss. Specifically, we adopt the widely used Learned Perceptual Image Patch Similarity (LPIPS) [ZIE*18] as a baseline for perceptual regularization. LPIPS computes the perceptual similarity between images by comparing feature maps extracted from a separately pretrained network. Following standard practice, we use a pretrained VGGNet [SZ15] as the backbone network for LPIPS in our experiments.

**Table 6:** *Ablation study comparing different structural regularization terms used in the discriminator energy.*

| Energy Term | FID (↓) | CLIP score (↑) |
|---|---|---|
| MSE | 12.11 | 0.3102 |
| MSE + LPIPS | 10.74 | 0.3213 |
| MSE + SCE (Ours) | **9.42** | **0.3428** |

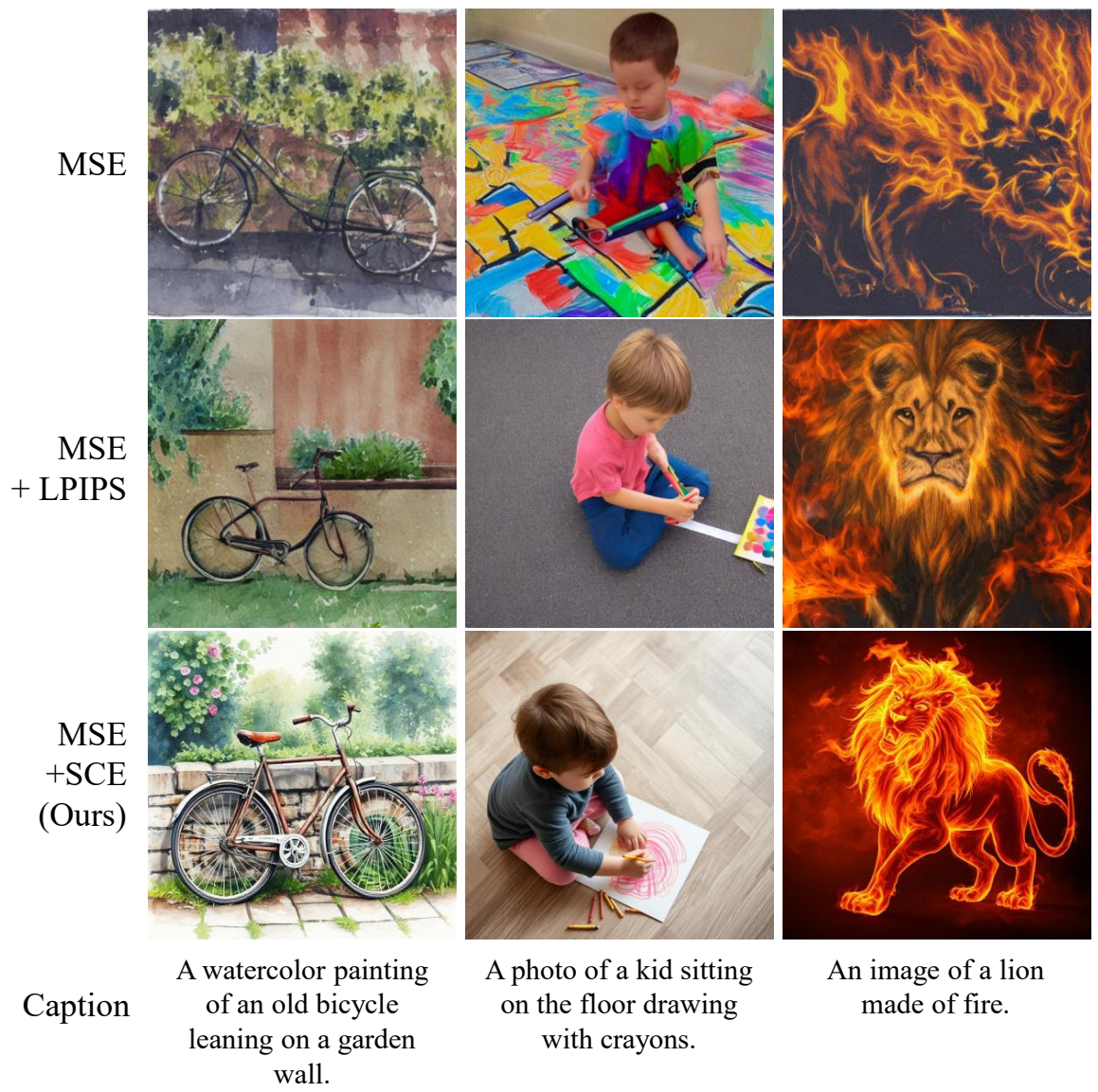

**Figure 7:** *Qualitative comparison of different structural loss configurations in the proposed framework.*

For all experiments, we used a fixed subset of the LAION-5B dataset [SBV*22] across different parameter settings. To ensure a fair comparison, all training configurations and hyperparameters other than the perceptual loss were fixed according to the setup described in Sec. 4.1. Following the text-to-image generation evaluation protocol in Sec. 4.2.1, we randomly sampled 20K image-text pairs from the COCO2014 validation set [LMB*14] and the Flickr30k dataset [YLHH14], and evaluated the quality of generation by measuring the FID and the CLIP score. To evaluate the impact of structural consistency energy on the performance of the proposed method, we conducted an ablation study using the following three configurations for the discriminator energy term: (1) using only the reconstruction-based MSE term (*i.e.*, $D(z_t) = D_{\text{MSE}}(z_t)$), (2) using both the MSE and our proposed structural consistency energy, and (3) using MSE and LPIPS [ZIE*18] as a conventional perceptual loss. The comparison results are summarized in Table 6.

We observe that the configuration using the proposed struc-tural consistency energy achieves the best performance in both the FID and the CLIP score. This demonstrates that directly enforcing multi-scale feature alignment between the encoder and decoder within the VAE structure leads to more effective discriminator updates and improved training efficiency. Since our structural consistency energy is fully integrated into the latent diffusion model, it provides strong regularization without requiring any external network components. In contrast, LPIPS evaluates perceptual similarity based on feature activations extracted from a separately pretrained network (*e.g.*, VGGNet), which is independent of the discriminator's architecture. As a result, the LPIPS-based loss may not provide gradient signals that are optimally aligned with the discriminator's learning dynamics. Moreover, LPIPS introduces additional computational overhead due to external feature extraction during training.

These results highlight a key advantage of our approach: The structural consistency energy enables perceptual regularization that is effective and computationally efficient, as it leverages the internal structure of the model rather than relying on external priors. This design ensures that perceptual alignment contributes directly to adversarial and generative objectives, leading to higher-quality synthesis.

The qualitative results are depicted in Fig. 7. When using only the reconstruction-based MSE loss, without any perceptual constraint, such as structural consistency energy or LPIPS, we observe noticeable blurring artifacts in the generated images. This is because MSE loss merely minimizes the pixel-level difference between the input and output latent representations, without accounting for high-level structural attributes. As a result, the generator tends to produce oversmoothed outputs that lack perceptual detail. Introducing LPIPS as a perceptual loss helps mitigate this blurring to some extent, leading to improved texture and perceptual realism. However, compared to our proposed structural consistency energy, the output still exhibits inferior sharpness and reduced sample diversity. This is likely due to the fact that LPIPS evaluates similarity based on features extracted from a separately pretrained network, which is not aligned with the internal structure of the diffusion model's discriminator. Furthermore, LPIPS typically operates on a single layer of the pretrained network, making it less effective in preserving multi-scale structural information. In contrast, the proposed SCE delivers the most perceptually faithful and structurally coherent results. By enforcing direct feature alignment between the encoder and decoder at multiple scales within the VAE structure, SCE allows the discriminator to guide the generator with more informative and structurally aware feedback. This leads to sharper, more diverse outputs that better capture the complex data distribution. Additionally, since SCE leverages the pretrained VAE prior, it facilitates stable training even on diverse and large-scale datasets, further supporting robust generative performance.

## 5 Conclusion

We introduced Latent Diffusion Generative Adversarial Networks (LD-GAN), a novel framework that seamlessly integrates adversarial learning into latent diffusion models without requiring an extra discriminator network. LD-GAN utilizes the pretrained VAE encoder and decoder as an energy-based discriminator for latent diffusion, allowing adversarial training in the structured latent space. This improves perceptual sharpness and sample fidelity while preserving the original diffusion model pipeline and minimizing computational overhead. Furthermore, we introduced a structural consistency energy to align encoder and decoder feature representations, stabilizing training and reinforcing structural coherence without external pretrained networks. Extensive experiments demonstrate that LD-GAN consistently improves generation quality across diverse tasks, including conditional text-to-image and 2D-to-3D generations. While LD-GAN eliminates the need for a separate discriminator, it requires decoder fine-tuning during adversarial training, introducing an additional optimization component. Moreover, balancing the adversarial and denoising objectives remains crucial for maintaining consistency with the pretrained latent prior. In future work, we aim to develop strategies to balance these objectives and extend LD-GAN to multi-modal or cross-domain settings.

## References

[BNH*22] Balaji, Yogesh, Nah, Seungjun, Huang, Xun, et al. "EDIFF-I: Text-to-Image Diffusion Models with an Ensemble of Expert Denoisers". *arXiv preprint arXiv:2211.01324* (2022) 8.

[DDS*09] Deng, Jia, Dong, Wei, Socher, Richard, et al. "ImageNet: A Large-Scale Hierarchical Image Database". *Proceedings of the IEEE/CVF Conference on Computer Vision and Pattern Recognition (CVPR)*. 2009, 248–255 19, 20.

[DFK*22] Downs, Laura, Francis, Anthony, Koenig, Nate, et al. "Google Scanned Objects: A High-Quality Dataset of 3D Scanned Household Items". *Proceedings of the International Conference on Robotics and Automation (ICRA)*. 2022 13.

[DM19] Du, Yilun and Mordatch, Igor. "Implicit Generation and Modeling with Energy-Based Models". *Proceedings of the Neural Information Processing Systems (NeurIPS)*. 2019 2, 3, 5.

[DN21] Dhariwal, Prafulla and Nichol, Alex. "Diffusion models beat GANs on image synthesis". *Proceedings of the Advances in Neural Information Processing Systems (NeurIPS)*. 2021, 8780–8794 1–3.

[DSS*23] Deitke, Matt, Schwenk, Dustin, Salvador, Jordi, et al. "Objaverse: A Universe of Annotated 3D Objects". *Proceedings of the IEEE/CVF Conference on Computer Vision and Pattern Recognition (CVPR)*. 2023 12.

[GAA*22] Gal, Rinon, Alaluf, Yuval, Atzmon, Yuval, et al. "An Image Is Worth One Word: Personalizing Text-to-Image Generation Using Textual Inversion". *arXiv preprint arXiv:2208.01618* (2022) 6–9, 11.

[GPM*14] Goodfellow, Ian, Pouget-Abadie, Jean, Mirza, Mehdi, et al. "Generative adversarial nets". *Proceedings of the Advances in Neural Information Processing Systems (NeurIPS)*. Vol. 27. 2014, 2672–2680 1–3.

[GPM*22] Gal, Rinon, Patashnik, Or, Maron, Haggai, et al. "StyleGAN-NADA: CLIP-Guided Domain Adaptation of Image Generators". *ACM Transactions on Graphics (ToG)* 41.4 (2022), 1–13 9.

[GWJ*20] Grathwohl, Will, Wang, Kuan-Chieh, Jacobsen, Joern-Henrik, et al. "Your classifier is secretly an energy based model and you should treat it like one". *Proceedings of the International Conference on Learning Representations (ICLR)*. 2020 7.

[HJA20] Ho, Jonathan, Jain, Ajay, and Abbeel, Pieter. "Denoising diffusion probabilistic models". *Proceedings of the Advances in Neural Information Processing Systems (NeurIPS)*. Vol. 33. 2020, 6840–6851 1–3, 6.

[HRS16] Hardt, Moritz, Recht, Ben, and Singer, Yoram. "Train Faster, Generalize Better: Stability of Stochastic Gradient Descent". *Proceedings of the International Conference on Machine Learning (ICML)*. Vol. 48. PMLR, 2016, 1225–1234 6.

[HRU*17] Heusel, Martin, Ramsauer, Hubert, Unterthiner, Thomas, et al. "GANs Trained by a Two Time-Scale Update Rule Converge to a Local Nash Equilibrium". *Proceedings of the Advances in Neural Information Processing Systems (NeurIPS)*. 2017, 6626–6637 8, 9, 14, 20.

[JKK*25] Jun, U, Ko, Jaeeun, Kang, Jiwoo, et al. "Generative adversarial diffusion". *Proceedings of the IEEE/CVF International Conference on Computer Vision*. 2025, 16786–16796 1.

[KAAL22] Karras, Tero, Aittala, Miika, Aila, Timo, and Laine, Samuli. "Elucidating the design space of diffusion-based generative models". *Proceedings of the Advances in Neural Information Processing Systems (NeurIPS)*. 2022 2.

[KLA19] Karras, Tero, Laine, Samuli, and Aila, Timo. "A style-based generator architecture for generative adversarial networks". *Proceedings of the IEEE/CVF Conference on Computer Vision and Pattern Recognition (CVPR)*. 2019, 4401–4410 20.

[KLL21] Kang, Jiwoo, Lee, Seongmin, and Lee, Sanghoon. "Competitive learning of facial fitting and synthesis using UV energy". *IEEE Transactions on Systems, Man, and Cybernetics: Systems* 52.5 (2021), 2858–2873 3.

[KW14] Kingma, Diederik P and Welling, Max. "Auto-encoding variational bayes". *Proceedings of the International Conference on Learning Representations (ICLR)*. 2014 3.

[KZB*24] Kang, Minguk, Zhang, Richard, Barnes, Connelly, et al. "Distilling diffusion models into conditional GANs". *Proceedings of the European Conference on Computer Vision (ECCV)*. 2024, 428–447 2, 3.

[LC98] Lehmann, Erich L. and Casella, George. *Theory of Point Estimation*. 2nd. New York: Springer, 1998 7.

[LGL*24] Long, Xiaoxiao, Guo, Yuan-Chen, Lin, Cheng, et al. "Wonder3D: Single Image to 3D Using Cross-Domain Diffusion". *Proceedings of the IEEE Conference on Computer Vision and Pattern Recognition (CVPR)*. 2024, 9970–9980 11.

[LLL*24] Li, Peng, Liu, Yuan, Long, Xiaoxiao, et al. "Era3D: High-Resolution Multiview Diffusion Using Efficient Row-Wise Attention". *Proceedings of the Advances in Neural Information Processing Systems (NeurIPS)*. Vol. 37. 2024, 55975–56000 6–8, 10–12.

[LLW*23] Li, Yifan, Liu, Huan, Wu, Qian, et al. "GLIGEN: Open-Set Grounded Text-to-Image Generation". *Proceedings of the IEEE/CVF Conference on Computer Vision and Pattern Recognition (CVPR)*. 2023, 22511–22521 6–10.

[LLWT15] Liu, Ziwei, Luo, Ping, Wang, Xiaogang, and Tang, Xiaoou. "Deep learning face attributes in the wild". *Proceedings of the IEEE International Conference on Computer Vision (ICCV)*. 2015, 3730–3738 20.

[LLZ*23] Liu, Yuan, Lin, Cheng, Zeng, Zijiao, et al. "SyncDreamer: Generating Multiview-Consistent Images from a Single-View Image". *arXiv preprint arXiv:2309.03453* (2023) 6–8, 10–12.

[LMB*14] Lin, Tsung-Yi, Maire, Michael, Belongie, Serge, et al. "Microsoft COCO: Common Objects in Context". *Proceedings of the European Conference on Computer Vision (ECCV)*. 2014, 740–755 8, 13, 14, 18, 19.

[LWV*23] Liu, Ruoshi, Wu, Rundi, Van Hoorick, Basile, et al. "Zero-1-to-3: Zero-Shot One Image to 3D Object". *Proceedings of the IEEE/CVF International Conference on Computer Vision (ICCV)*. 2023, 9298–9309 11.

[LZB*25] Lu, Cheng, Zhou, Yuhao, Bao, Fan, et al. "DPM-Solver++: Fast solver for guided sampling of diffusion probabilistic models". *Machine Intelligence Research* 22 (2025), 730–751 3.

[LZZ*22a] Li, Liunian Harold, Zhang, Pengchuan, Zhang, Haotian, et al. "Grounded Language-Image Pre-Training". *Proceedings of the IEEE/CVF Conference on Computer Vision and Pattern Recognition (CVPR)*. 2022, 10965–10975 9, 10.

[LZZ*22b] Li, Liunian Harold, Zhang, Pengchuan, Zhang, Haotian, et al. "Grounded Language-Image Pre-Training". *Proceedings of the IEEE/CVF Conference on Computer Vision and Pattern Recognition (CVPR)*. 2022, 10965–10975 9.

[ND21] Nichol, Alexander Quinn and Dhariwal, Prafulla. "Improved denoising diffusion probabilistic models". *Proceedings of the International Conference on Machine Learning (ICML)*. PMLR, 2021 2.

[NGH*22] Nie, Weili, Guo, Brandon, Huang, Yujia, et al. "Diffusion models for adversarial purification". *Proceedings of the International Conference on Machine Learning (ICML)*. PMLR, 2022 1, 2.

[OKB11] Ordonez, Vicente, Kulkarni, Girish, and Berg, Tamara L. "Im2Text: Describing Images Using 1 Million Captioned Photographs". *Proceedings of the Advances in Neural Information Processing Systems (NeurIPS)*. 2011 9.

[PEL*24] Podell, Dustin, English, Zion, Lacey, Kyle, et al. "SDXL: Improving Latent Diffusion Models for High-resolution Image Synthesis". *Proceedings of the International Conference on Learning Representations (ICLR)*. 2024 8.

[PHN*20] Pang, Bo, Han, Tian, Nijkamp, Erik, et al. "Learning latent space energy-based prior model". *Proceedings of the Advances in Neural Information Processing Systems (NeurIPS)*. Vol. 33. 2020, 21994–22008 3.

[RBL*22] ROMBACH, ROBIN, BLATTMANN, ANDREAS, LORENZ, DOMINIK, et al. "High-resolution image synthesis with latent diffusion models". *Proceedings of the IEEE/CVF Conference on Computer Vision and Pattern Recognition (CVPR)*. 2022, 10684–10695 1, 2, 6–8, 13, 19, 20.

[RKH*21] RADFORD, ALEC, KIM, JONG WOOK, HALLACY, CHRIS, et al. "Learning Transferable Visual Models from Natural Language Supervision". *Proceedings of the International Conference on Machine Learning (ICML)*. PMLR, 2021, 8748–8763 8, 9, 14.

[SBV*22] SCHUHMANN, CHRISTOPH, BEAUMONT, ROMAIN, VENCU, RICHARD, et al. "LAION-5B: An Open Large-Scale Dataset for Training Next Generation Image-Text Models". *Proceedings of the Advances in Neural Information Processing Systems (NeurIPS)*. Vol. 35. 2022, 25278–25294 7, 8, 13, 14, 18–20.

[SE19] SONG, YANG and ERMON, STEFANO. "Generative modeling by estimating gradients of the data distribution". *Proceedings of the Advances in Neural Information Processing Systems (NeurIPS)*. Vol. 32. 2019, 11918–11930 2.

[SGZ*16] SALIMANS, TIM, GOODFELLOW, IAN, ZAREMBA, WOJCIECH, et al. "Improved techniques for training GANs". *Proceedings of the Neural Information Processing Systems (NeurIPS)*. 2016, 2234–2242 5.

[SHCS22] SALIMANS, TIM, HO, JONATHAN, CHEN, XI, and SOHL-DICKSTEIN, JASCHA. "Progressive distillation for fast sampling of diffusion models". *Proceedings of the International Conference on Learning Representations (ICLR)*. 2022 1, 2.

[SLZ*19] SHAO, SHUANG, LI, ZHI, ZHANG, TIANLONG, et al. "Objects365: A Large-Scale, High-Quality Dataset for Object Detection". *Proceedings of the IEEE/CVF International Conference on Computer Vision (ICCV)*. 2019, 8430–8439 9.

[SME21] SONG, JIAMING, MENG, CHENLIN, and ERMON, STEFANO. "Denoising diffusion implicit models". *Proceedings of the International Conference on Learning Representations (ICLR)*. 2021 2, 3.

[SPH*23] SINGER, URIEL, POLYAK, ADAM, HAYES, THOMAS, et al. "Make-a-video: Text-to-video generation without text-video data". *Proceedings of the International Conference on Learning Representations (ICLR)*. 2023 1.

[SWCM21] SEHWAG, VIKASH, WANG, SHIQI, CHIANG, MUNG, and MITTAL, PRATEEK. "SSD: A Unified Framework for Self-Supervised Outlier Detection". *Proceedings of the International Conference on Learning Representations (ICLR)*. 2021 2, 18, 20.

[SWMG15] SOHL-DICKSTEIN, JASCHA, WEISS, ERIC, MAHESWARANATHAN, NIRU, and GANGULI, SURYA. "Deep unsupervised learning using nonequilibrium thermodynamics". *Proceedings of the International Conference on Machine Learning (ICML)*. PMLR, 2015, 2256–2265 1.

[SZ15] SIMONYAN, KAREN and ZISSERMAN, ANDREW. "Very Deep Convolutional Networks for Large-Scale Image Recognition". *Proceedings of the International Conference on Learning Representations (ICLR)*. 2015 14.

[TH24] TRINH, LUAN THANH and HAMAGAMI, TOMOKI. "Latent denoising diffusion gan: Faster sampling, higher image quality". *IEEE Access* 12 (2024), 78161–78172 20.

[TMJS20] TACK, JIHOON, MO, SANGWOO, JEONG, JONGHEON, and SHIN, JINWOO. "CSI: Novelty Detection via Contrastive Learning on Distributionally Shifted Instances". *Proceedings of the Neural Information Processing Systems (NeurIPS)*. 2020 2.

[UVL16] ULYANOV, DMITRY, VEDALDI, ANDREA, and LEMPITSKY, VICTOR. "Instance normalization: The missing ingredient for fast stylization". *arXiv preprint arXiv:1607.08022* (2016) 4, 13.

[VKK21] VAHDAT, ARASH, KREIS, KARSTEN, and KAUTZ, JAN. "Score-based generative modeling in latent space". *Proceedings of the Advances in Neural Information Processing Systems (NeurIPS)*. Vol. 34. 2021, 11287–11302 1–3.

[WBSS04] WANG, ZHOU, BOVIK, ALAN C., SHEIKH, HAMID R., and SIMONCELLI, EERO P. "Image Quality Assessment: From Error Visibility to Structural Similarity". *IEEE Transactions on Image Processing (TIP)* 13.4 (2004), 600–612 11.

[WCHN22] WATSON, DANIEL, CHAN, WILLIAM, HO, JONATHAN, and NOROUZI, MOHAMMAD. "Learning fast samplers for diffusion models by differentiating through sample quality". *Proceedings of the International Conference on Learning Representations (ICLR)*. 2022 3.

[WH18] WU, YUXIN and HE, KAIMING. "Group normalization". *Proceedings of the European Conference on Computer Vision (ECCV)*. 2018, 3–19 13.

[WPD*23] WANG, ZEKAI, PANG, TIANYU, DU, CHAO, et al. "Better diffusion models further improve adversarial training". *Proceedings of the International Conference on Machine Learning (ICML)*. PMLR, 2023 2.

[WZF*23] WU, TONG, ZHANG, JIARUI, FU, XIAO, et al. "OmniObject3D: Large-Vocabulary 3D Object Dataset for Realistic Perception, Reconstruction and Generation". *Proceedings of the IEEE/CVF Conference on Computer Vision and Pattern Recognition (CVPR)*. 2023 10, 13.

[WZH*23] WANG, ZHENDONG, ZHENG, HUANGJIE, HE, PENGCHENG, et al. "Diffusion-GAN: Training GANs with diffusion". *Proceedings of the International Conference on Learning Representations (ICLR)*. 2023 1–3, 19, 20.

[XKV22] XIAO, ZHISHENG, KREIS, KARSTEN, and VAHDAT, ARASH. "Tackling the Generative Learning Trilemma with Denoising Diffusion GANs". *Proceedings of the International Conference on Learning Representations (ICLR)*. 2022 2.

[XZL21] XIE, JIANWEN, ZHENG, ZILONG, and LI, PING. "Learning energy-based model with variational auto-encoder as amortized sampler". *Proceedings of the AAAI Conference on Artificial Intelligence (AAAI)*. Vol. 35. 12. 2021, 10441–10451 3.

[XZXH24] XU, YANWU, ZHAO, YANG, XIAO, ZHISHENG, and HOU, TINGBO. "UFOGen: You Forward Once Large Scale Text-to-Image Generation via Diffusion GANs". *Proceedings of the IEEE Conference on Computer Vision and Pattern Recognition (CVPR)*. 2024, 8196–8206 2.

[YGZ*24] YIN, TIANWEI, GHARBI, MICHAËL, ZHANG, RICHARD, et al. "One-step diffusion with distribution matching distillation". *Proceedings of the IEEE/CVF Conference on Computer Vision and Pattern Recognition (CVPR)*. 2024, 6613–6623 2, 3.

[YLHH14] YOUNG, PETER, LAI, ALICE, HODOSH, MICAH, and HOCKENMAIER, JULIA. "From Image Descriptions to Visual Denotations: New Similarity Metrics for Semantic Inference over Event Descriptions". *Transactions of the Association for Computational Linguistics* 2 (2014), 67–78 8, 9, 13, 14, 18, 19.

[YSZ*15] YU, FISHER, SEFF, ARI, ZHANG, YINDA, et al. "LSUN: Construction of a large-scale image dataset using deep learning with humans in the loop". *arXiv preprint arXiv:1506.03365* (2015) 20.

[ZIE*18] ZHANG, RICHARD, ISOLA, PHILLIP, EFROS, ALEXEI A., et al. "The Unreasonable Effectiveness of Deep Features as a Perceptual Metric". *Proceedings of the IEEE/CVF Conference on Computer Vision and Pattern Recognition (CVPR)*. 2018 2, 3, 5, 6, 11, 14.

[ZML17] ZHAO, JUNBO, MATHIEU, MICHAEL, and LECUN, YANN. "Energy-based generative adversarial network". *Proceedings of the International Conference on Learning Representations (ICLR)*. 2017 1–4, 6, 7.

[ZRA23] ZHANG, LVMIN, RAO, ANYI, and AGRAWALA, MANEESH. "Adding conditional control to text-to-image diffusion models". *Proceedings of the IEEE International Conference on Computer Vision (ICCV)*. 2023 8, 20.

## A Hyperparameters Analysis

In this section, we analyzed the sensitivity of the proposed method with respect to various hyperparameters, such as the adversarial balancing parameter $\lambda_{adv}$ in (8), the structural consistency energy balancing parameter $\lambda_{SC}$ in (10), and the number of feature layers $|\mathcal{L}|$ in (9). Specifically, in Sec. A.1, we analyzed the sensitivity of the proposed method to the adversarial weight $\lambda_{adv}$. In addition, we investigated the sensitivity of the proposed method to the number of feature layers $|\mathcal{L}|$ on structural consistency and generation quality in Sec. A.2. Furthermore, in Sec. A.3, we examined the impact of the balancing parameter $\lambda_{SC}$ of the structural consistency energy on the dynamics of training.

### A.1 Sensitivity Analysis of $\lambda_{adv}$

In this section, we analyzed the sensitivity of the proposed method with respect to the adversarial weight $\lambda_{adv}$ in (8), using a fixed subset of the LAION-5B dataset [SBV*22] across different parameter settings. To ensure a fair comparison, other hyperparameters except $\lambda_{adv}$ were fixed to the values specified in Sec. 4.1. Following the same evaluation protocol as in Sec. 4.2.1, we randomly sampled 20K image-text pairs from the COCO2014 validation set [LMB*14] and the Flickr30k dataset [YLHH14], and evaluated the quality of the generation using both the FID and the CLIP score.

We conducted experiments using three values of $\lambda_{adv}$: 0.05, 0.01, and 0.005. The performance comparison is summarized in Table 7. It is shown that $\lambda_{adv} = 0.01$ gives the best empirical performance in text-to-image generation. When $\lambda_{adv}$ is set too high (*i.e.*, 0.05), the generator becomes excessively influenced by the adversarial signal, leading to reduced diversity and the appearance of minor artifacts due to the discriminator dominating the training dynamics. In contrast, when $\lambda_{adv}$ is set too low (*i.e.*, 0.005), the adversarial guidance becomes negligible, resulting in less perceptual sharpness and limited improvement over the baseline. These findings suggest that an appropriately balanced adversarial contribution is essential to maintain both fidelity and diversity in the generated outputs.

Furthermore, we observed consistent performance patterns across multiple generation tasks, including conditional text-to-image generation and 2D-to-3D generation. In all cases, $\lambda_{adv} = 0.01$ provided competitive performance without requiring additional tuning. We attribute this robustness to the fact that the adversarial loss in our method operates within the structured latent space of the diffusion model, which helps stabilize gradient propagation. This observation suggests that a single value of $\lambda_{adv}$ generalizes well across diverse tasks, alleviating the need for a task-specific reconfiguration of hyperparameters.

**Table 7:** *Sensitivity analysis of the adversarial weight $\lambda_{adv}$ in the proposed method.*

| $\lambda_{adv}$ | FID (↓) | CLIP score (↑) |
|---|---|---|
| 0.05 | 10.97 | 0.3212 |
| 0.01 | **9.42** | **0.3428** |
| 0.005 | 12.31 | 0.3187 |

**Table 8:** *Analysis according to the number and configuration of feature layer pairs $\mathcal{L}$ used in computing the structural consistency energy.*

| $\lvert\mathcal{L}\rvert$ | $\mathcal{L}$ | FID (↓) | CLIP score (↑) |
|---|---|---|---|
| 0 | $\emptyset$ | 12.11 | 0.3102 |
| 1 | $\{1\}$ | 11.19 | 0.3114 |
| | $\{2\}$ | 10.37 | 0.3255 |
| | $\{3\}$ | 10.86 | 0.3178 |
| 2 | $\{1,2\}$ | 10.12 | 0.3175 |
| | $\{2,3\}$ | 9.87 | 0.3284 |
| | $\{1,3\}$ | 10.75 | 0.3271 |
| 3 | $\{1,2,3\}$ | **9.42** | **0.3428** |

### A.2 The Configuration of Feature Layers $\mathcal{L}$

In this section, we analyzed the sensitivity of the proposed method to the configuration of feature layers $\mathcal{L}$ used in the structural consistency energy defined in (9). We used a fixed subset of the LAION-5B dataset [SBV*22] for training, and all other hyperparameters are fixed to the values specified in Sec. 4.1 to ensure a fair comparison. Following the same evaluation protocol as in Sec. 4.2.1, we randomly sampled 20K image-text pairs from the COCO2014 validation set [LMB*14] and the Flickr30k dataset [YLHH14]. We evaluated the quality of generation by measuring both the FID and the CLIP score.

Our method is implemented using the Stable Diffusion [SWCM21] architecture as the backbone. The VAE of Stable Diffusion consists of an encoder with three downsampling stages and a decoder that includes three corresponding upsampling stages. We defined each feature layer pair in $\mathcal{L}$ as the encoder feature immediately before downsampling and the corresponding decoder feature immediately after the aligned upsampling stage. Hence, if $n$ denotes the index of the stage, then $\mathcal{L} = \{n\}$ refers to the feature pair at that level. The total number of feature layer pairs used in computing the structural consistency energy is denoted by $|\mathcal{L}|$.

To assess the sensitivity to this design choice, we conduct experiments with all combinations of $|\mathcal{L}| = 0,1,2,3$, corresponding to $\mathcal{L} \in \{\emptyset, \{1\}, \{2\}, \{3\}, \{1,2\}, \{1,3\}, \{2,3\}, \{1,2,3\}\}$. Here, $\mathcal{L} = \emptyset$ corresponds to the case where the structural consistency energy is not used, *i.e.*, $\lambda_{SC} = 0$. This serves as a baseline to isolate the impact of the structural consistency term from the rest of the training objectives. In preliminary tests, we observed that the specific location of the feature pair (within the same resolution scale) has minimal impact on performance, while the deepest layer ($\{3\}$) consistently demonstrates the best results. Therefore, we used this as the basis for further combinations of layers.

Table 8 reports the results for various values of $|\mathcal{L}|$. We observe that utilizing the three pairs of feature layers ($|\mathcal{L}| = 3$) yields the best empirical performance for text-to-image generation. Additionally, we observe a consistent improvement in generation quality as $|\mathcal{L}|$ increases, under the constraint of using each representative feature pair at each resolution level. This can be attributed to the en-hanced ability of the model to preserve multi-scale structural information when low, mid, and high-level features are jointly aligned. This highlights the importance of aligning intermediate features at multiple scales to reinforce structural consistency and perceptual quality. By preventing the model from overfitting to a specific resolution level, the structural consistency energy helps the generator better capture the structural attributes of images.

**Table 9:** *Sensitivity analysis of the structural consistency energy weight $\lambda_{SC}$ on generation performance.*

| $\lambda_{SC}$ | FID (↓) | CLIP score (↑) |
|---|---|---|
| 0.1 | 10.12 | 0.3246 |
| 0.05 | **9.42** | **0.3428** |
| 0.01 | 9.87 | 0.3384 |

Furthermore, we observe similar trends across various generative tasks, including conditional text-to-image generation, and 2D-to-3D generation. These results suggest that incorporating multi-scale feature pairs improves performance across various generative tasks. We attribute this effect to the capacity of the structural consistency energy to align intermediate encoder and decoder features more effectively, thereby reinforcing both structured priors and perceptual fidelity. This cross-task consistency also highlights the robustness and generalizability of the proposed regularization scheme.

### A.3 Balancing Parameter $\lambda_{SC}$

In this section, we analyzed the sensitivity of the proposed method to the structural consistency energy by varying the balancing parameter $\lambda_{SC}$ defined in Eq. (10). We used a fixed subset of the LAION-5B dataset [SBV*22] across different parameter settings to ensure consistency in the training data. To ensure a fair comparison, all other hyperparameters except $\lambda_{SC}$ were fixed to the values specified in Sec. 4.1. Following the same evaluation protocol as in Sec. 4.2.1, we randomly sampled 20K image-text pairs from the COCO2014 validation set [LMB*14] and the Flickr30k dataset [YLHH14]. We evaluated the quality of generation by measuring both the FID and the CLIP score. We conducted experiments with three values of $\lambda_{SC}$ : 0.01, 0.05, and 0.1. The performance comparison results are presented in Table 9.

The results in Table 9 show that the optimal performance is attained when $\lambda_{SC} = 0.05$. When $\lambda_{SC}$ is too small (*i.e.*, 0.01), the model tends to focus primarily on minimizing the latent reconstruction error, and the influence of the structural consistency energy becomes negligible. As a result, the model may fail to capture high-level structural patterns, leading to blurred outputs, reduced perceptual sharpness, and diminished structural coherence. In contrast, when $\lambda_{SC}$ is excessively large (*i.e.*, 0.1), the model places undue emphasis on aligning encoder and decoder features, which may hinder its ability to accurately reconstruct latent representations. This can lead to an over-regularization effect, whereby the model emphasizes the preservation of structural alignment across scales but underfits to fine details such as color and texture. Moreover, strongly enforcing alignment in the feature space can restrict the model's ability to learn diverse representations, potentially reducing sample diversity and expressiveness. These results highlight the importance of balancing the contribution of structural consistency energy to ensure that the model maintains both high perceptual fidelity and diversity in the generated outputs.

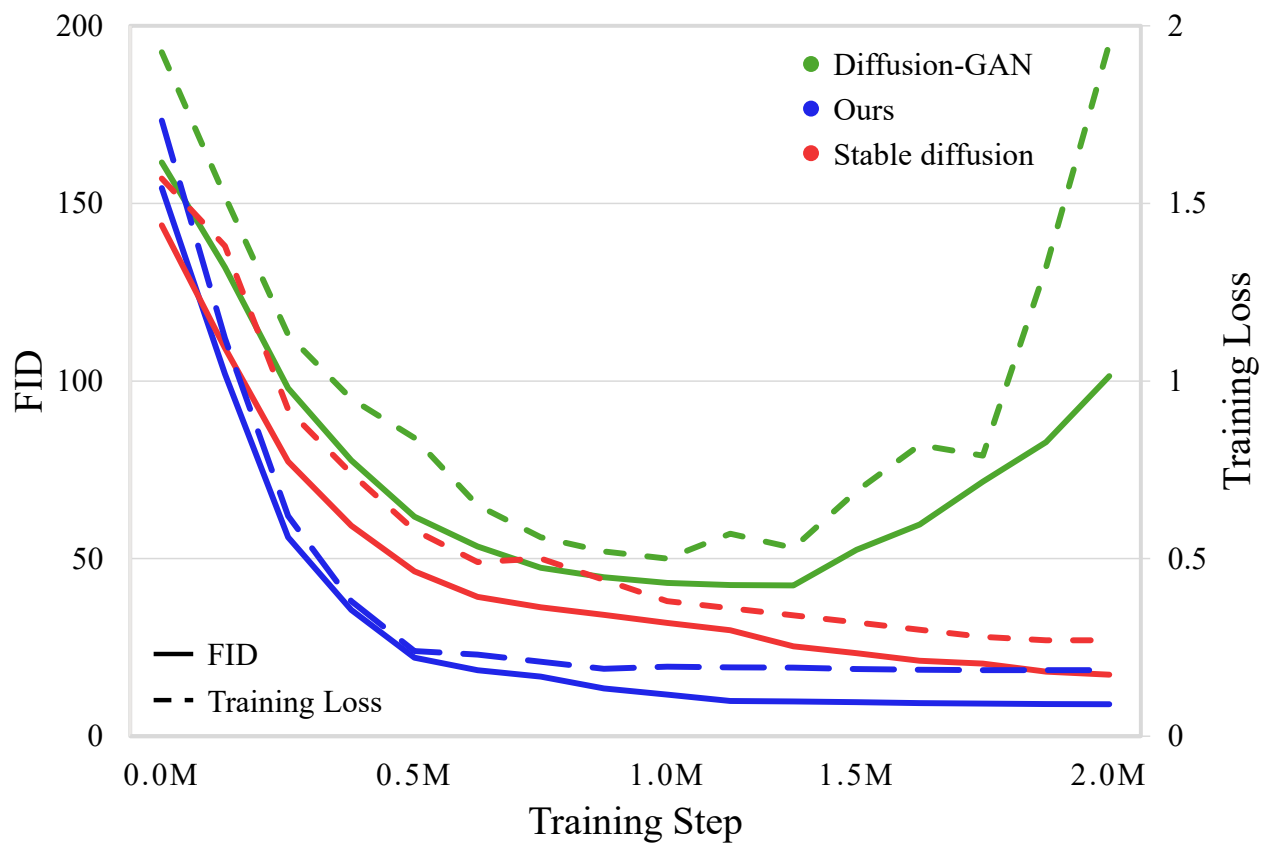

**Figure 8:** *Training convergence and stability comparison among the proposed LD-GAN, the GAN baseline (i.e., Diffusion-GAN [WZH*23]), and the diffusion baseline (i.e., Stable Diffusion [RBL*22]).*

Furthermore, similar trends were observed across various generative tasks, including conditional text-to-image generation, and 2D-to-3D generation. In our experiments, setting $\lambda_{SC} = 0.05$ consistently led to strong performance across various tasks, suggesting that this configuration serves as a robust default across multiple domains. While task-specific tuning may further optimize results in certain contexts, our findings indicate that the proposed structural consistency energy formulation generalizes effectively without necessitating extensive adjustments for each task.

## B Training Behavior and Efficiency Analysis

In this section, we evaluated the convergence speed and training stability of the proposed method in comparison with baseline approaches. As a diffusion model baseline, we adopted Stable Diffusion [RBL*22], which serves as the backbone in our experiments. For the baseline of adversarially trained diffusion, we used Diffusion-GAN [WZH*23], which applies GAN-based adversarial training directly to the latent denoising process. Specifically, Diffusion-GAN introduces an additional discriminator that operates during the denoising stage and is alternately trained with the generator (*i.e.*, UNet). In contrast, the proposed LD-GAN avoids introducing any additional networks by leveraging the pretrained encoder and decoder from the latent diffusion model to construct an energy-based discriminator. This design enables LD-GAN to utilize the structured prior learned from large-scale datasets, thereby reducing computational overhead while improving training stability.

For a fair comparison, all methods were trained from scratch on the same subset of the ImageNet dataset [DDS*09] under identical training schedules and parameter budgets. To ensure comparable model capacity, the Diffusion-GAN discrimina-tor was implemented using the same encoder architecture as the VAE, with an additional fully connected layer and sigmoid activation appended to the final feature layer. Following previous work [ZRA23; SWCM21], we measured the sample quality and convergence behavior by evaluating the Fréchet Inception Distance (FID) [HRU*17] at regular intervals during training. Additionally, we plotted training losses as learning curves to provide intuitive view of convergence. Figure 8 shows the learning curve and FID during the training process for the 2M steps.

**Comparison with Adversarial Diffusion Model.** Although Diffusion-GAN initially reduces FID and training loss, it begins to diverge after approximately 1.3M steps, indicating unstable training behavior. This instability likely stems from the binary classification-based discriminator objective, which struggles to capture fine-grained reconstruction errors and the full spectrum of sample quality in large-scale, multi-modal datasets. This led prior adversarial diffusion models, such as Diffusion-GAN [WZH*23] and LDDGAN [TH24], to be primarily evaluated on low-diversity datasets (*e.g.*, CelebA [LLWT15], LSUN [YSZ*15], FFHQ [KLA19]) with limited class or variation. As a result, adversarially trained diffusion models can become imbalanced or oscillatory, leading to degraded performance and potential mode collapses on large-scale, high-diversity and complex datasets such as LAION [SBV*22] and ImageNet [DDS*09] dataset.

**Convergence Behavior and Training Stability.** As shown in Fig. 8, the proposed LD-GAN demonstrates a consistently decreasing FID and training loss, maintains stable training throughout. This result is attributed to the use of an energy-based discriminator defined through reconstruction loss, which enables continuous and fine-grained evaluation of generated samples. Rather than merely classifying samples as real or fake, the discriminator offers fine-grained and informative feedback gradients that guide the generator to improve sample quality in a structured manner. Moreover, leveraging the pretrained VAE prior helps preserve distributional diversity and prevents collapse to specific modes, which is crucial for learning from complex datasets, such as ImageNet. Also, compared to the non-adversarial baseline (*i.e.*, Stable Diffusion), LD-GAN achieves faster convergence and lower final FID scores.

**Memory Overhead.** Regarding memory usage, in our mixed-precision (FP16, FP32) training environment, the baseline SD 2.1 U-Net generator requires nearly 9.7 GiB of memory. The proposed method, which additionally trains the VAE decoder as the discriminator, adds decoder weights, optimizer states, and activations, resulting in approximately 10.8 GiB of memory usage, which corresponds to nearly 11.8% increase compared to the baseline.

**Computational Complexity.** LD-GAN shares the same architecture as Stable Diffusion 2.1 [RBL*22], consisting of approximately 1B parameters: U-Net (∼865M), VAE decoder (∼83M), and encoder (∼80M). While Stable Diffusion only updates the U-Net during training, LD-GAN additionally trains the decoder as a discriminator, adding approximately 9.6% computational overhead per training step. Despite this additional cost, adversarial training leads to significantly faster convergence. As shown in Fig. 8, LD-GAN reaches an FID of 9.42 after approximately 1.2M steps, whereas non-adversarial baseline only achieves 13.79 after 2M steps. This corresponds to a reduction of nearly 40% in training steps and 34% in total training time, demonstrating that LD-GAN reduces overall computational cost while improving performance

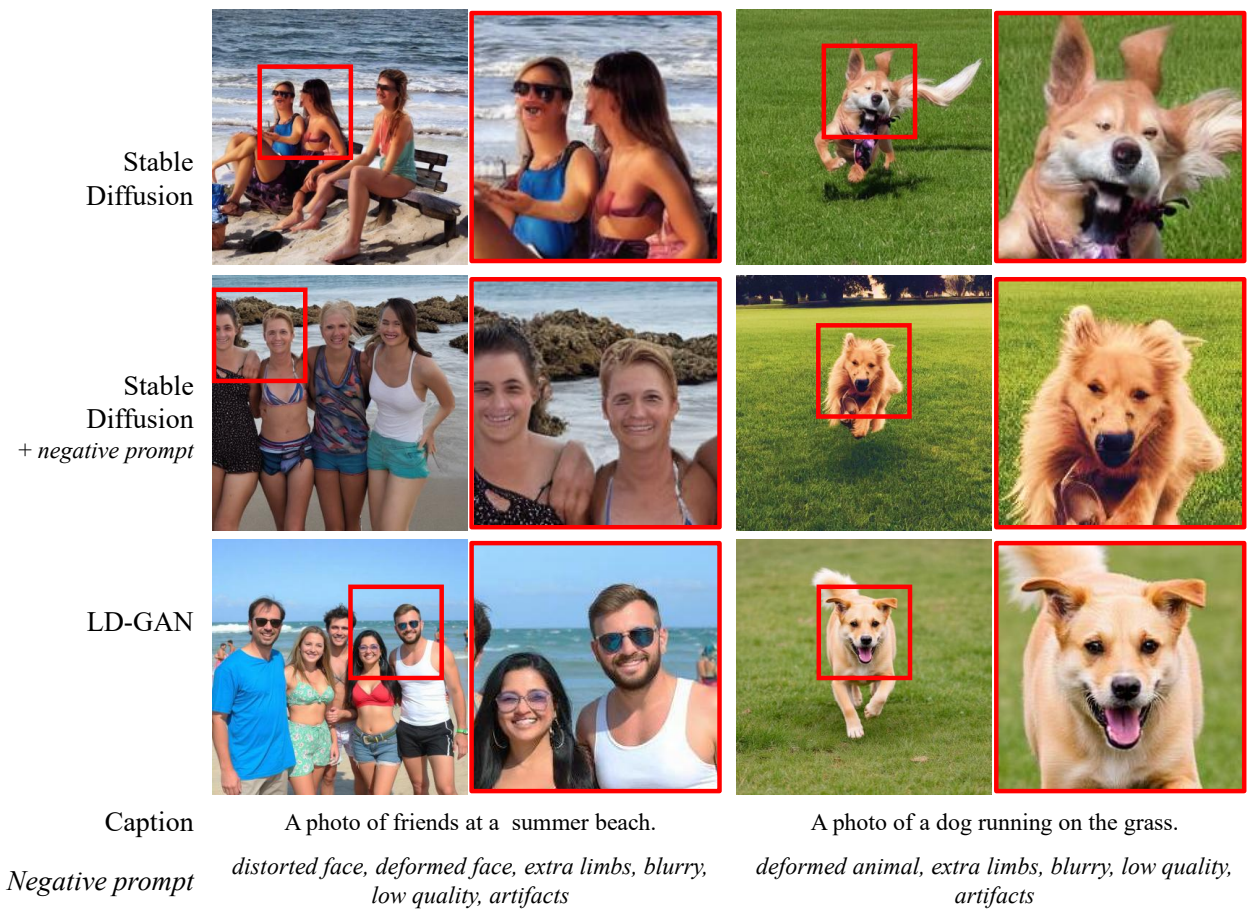

**Figure 9:** *Qualitative comparisons of local detail preservation under denoising score-matching baseline [RBL*22] and negative prompt settings.*

These results suggest that adversarial learning, when conducted in a structured and stable manner, facilitates more efficient training of the generator. In particular, the proposed energy-based discriminator with structural consistency energy encourages multi-scale alignment between encoder and decoder features, allowing the generator to more effectively capture the underlying data distribution and produce sharper, more coherent outputs.

## C Additional Qualitative Analysis

In this section, we provided additional qualitative analyses to examine the effect of the denoising objective and negative prompts on local detail preservation. As a baseline, we adopt Stable Diffusion [RBL*22], which is trained primarily with a denoising score-matching objective. We additionally evaluated the same baseline with negative prompts to analyze how prompt-level heuristics influence local artifacts. To ensure a fair comparison, all training configurations and hyperparameters are fixed according to the setup described in Sec. 4.1, and all models are trained on the same LAION-5B dataset [SBV*22]. The qualitative results are shown in Fig. 9.

**Denoising Objective Analysis.** Results obtained with the baseline, which relies on pure denoising score matching, exhibit locally attenuated high-frequency details, such as smoothed facial boundaries and fine textures, leading to slightly blurred local regions. This can be attributed to the L2 noise-matching objective, which tends to encourage averaging over ambiguous high-frequency components during training. In contrast, the proposed adversarial energy provides a structural constraint that counteracts the smoothing bias of L2 noise-matching, thereby improving high-frequency detail preservation.

**Negative Prompt Discussion.** As shown in Fig. 9, negative prompts reduce some locally distorted regions and mitigate certain local artifacts, such as distorted facial boundaries in human and irregular shapes around the ears and snout in dog, compared to the baseline. However, such improvements are prompt-dependent and rely on heuristic prompt engineering, whereas the proposed method improves the generation quality, including local sharpness and high-frequency detail, through adversarial training without modifying the prompt.